\documentclass[twoside,11pt]{article}

\usepackage[preprint]{jmlr2e}

\usepackage{natbib,hyperref} 
\let\cite\citep

\usepackage{algorithm}
\usepackage[noend]{algpseudocode}
\usepackage{amsmath}
\usepackage{amsthm,dsfont}
\theoremstyle{definition}
\usepackage[utf8]{inputenc} 
\usepackage{graphicx}
\usepackage{multirow}
\usepackage{microtype}
\usepackage{subfig}
\usepackage{booktabs}
\usepackage{enumitem}%
\usepackage{blkarray}

\usepackage[colorinlistoftodos,prependcaption,textsize=small]{todonotes}

\newtheorem{theorem}{Theorem}
\newtheorem{definition}{Definition}

\newtheorem{lemma}[theorem]{Lemma}
\newtheorem{proposition}[theorem]{Proposition}
\newtheorem{example}{Example}[section]

\usepackage{tikz}
\usetikzlibrary{automata}
\usetikzlibrary{bayesnet,arrows}
\usetikzlibrary{shapes,fit,calc,positioning}
\tikzset{box/.style={draw, diamond, thick, text centered, minimum height=0.5cm, minimum width=1cm, text width=0.9cm}}
\tikzset{line/.style={draw, thick, -latex'}}
\allowdisplaybreaks
\usepackage{pgfplots}

\newcommand{\framework}{\ensuremath{\mathsf{FairXplainer}}}

\newcommand{\numnonsensitive}{\ensuremath{m_1}}
\newcommand{\numsensitive}{\ensuremath{m_2}}
\newcommand{\numfeatures}{\ensuremath{m}}
\newcommand{\nonsensitive}{\ensuremath{\mathbf{X}}}
\newcommand{\sensitive}{\ensuremath{\mathbf{A}}}
\newcommand{\feature}{\ensuremath{\mathbf{Z}}}

\newcommand{\function}{\ensuremath{g}}

\newcommand{\alg}{\mathcal{M}}

\definecolor{affirmative}{rgb}{0.2,1,0.3}

\DeclareMathOperator*{\argmax}{arg\,max}
\DeclareMathOperator*{\argmin}{arg\,min}

\usepackage{comment}
\specialcomment{comment}{\begingroup\color{gray}}{\endgroup}
\excludecomment{comment}  %

\makeatletter
\newtheorem*{rep@theorem}{\rep@title}
\newcommand{\newreptheorem}[2]{%
	\newenvironment{rep#1}[1]{%
		\def\rep@title{\textbf{#2} \ref{##1}}%
		\begin{rep@theorem}}%
		{\end{rep@theorem}}}
\makeatother
\newreptheorem{theorem}{Theorem}
\newreptheorem{example}{Example}
\newreptheorem{proposition}{Proposition}

\title{``How Biased are Your Features?'': Computing Fairness Influence Functions with Global Sensitivity Analysis}

\author{\name Bishwamittra Ghosh \\
	\addr School of Computing\\
	National University of Singapore\\
	Singapore
	\AND
	\name Debabrota Basu \\
	\addr \'Equipe Scool, Univ. Lille, Inria, UMR 9189 - CRIStAL, CNRS\\
	Centrale Lille, France 
	\AND
	\name Kuldeep S. Meel  \\
	\addr School of Computing\\
	National University of Singapore\\
	Singapore}

\graphicspath{{../}}

\begin{document}

\maketitle
\begin{abstract}
	Fairness in machine learning has attained significant focus due to the widespread application in high-stake decision-making tasks. Unregulated machine learning classifiers can exhibit bias towards certain demographic groups in data, thus the quantification and mitigation of classifier bias is a central concern in fairness in machine learning.  In this paper, \textit{we aim to quantify the influence of different features in a dataset on the bias of a classifier}. To do this, we introduce the \textit{Fairness Influence Function} (FIF). This function breaks down bias into its components among individual features and the intersection of multiple features. The key idea is to represent existing group fairness metrics as the difference of the scaled conditional variances in the classifier's prediction and apply a decomposition of variance according to global sensitivity  analysis. To estimate FIFs, we instantiate an algorithm {\framework} that applies variance decomposition of classifier's prediction following local regression. 	Experiments demonstrate that {\framework} captures FIFs of individual feature and intersectional features, provides a better approximation of bias based on FIFs, demonstrates higher correlation of FIFs with fairness interventions, and detects changes in bias due to fairness affirmative/punitive actions in the classifier. 
	
	The code is available at \url{https://github.com/ReAILe/bias-explainer}. The extended version of the paper is at \url{https://arxiv.org/pdf/2206.00667.pdf}.
\end{abstract}
\section{Introduction}

The past few decades have witnessed a remarkable advancement in the field of machine learning, with its applications encompassing crucial decision-making processes such as college admissions~\cite{martinez2021using}, recidivism prediction~\cite{tollenaar2013method}, job applications~\cite{ajunwa2016hiring}, and more.  In such applications, the deployed machine learning classifier often demonstrates bias towards certain demographic groups present in the data~\cite{dwork2012fairness}. For instance, a machine learning classifier employed in college admissions may show a disproportionate preference for White-male candidates over Black-female candidates. This can occur due to historical biases present in the admission data, the classifier's accuracy-focused learning objective, or a combination of the two~\cite{landy1978correlates,zliobaite2015relation,berk2019accuracy}. Following such phenomena, multiple group-based fairness metrics, such as \textit{statistical parity}, \textit{equalized odds}, \textit{predictive parity}, etc.\ have been proposed to quantify the bias of the classifier on a dataset.  For example, a statistical parity of $ 0.6 $ for the college admission classifier would indicate that White-male candidates are offered admission $ 60\% $ more frequently than Black-female candidates~\cite{garg2020fairness,besse2021survey}.

Fairness metrics measure global bias, but do not detect or explain its sources~\cite{begley2020explainability,lundberg2020explaining,pan2021explaining}.
In order to diagnose the emergence of bias in the predictions of a classifier, it is important to compute explanations, such as how different features attribute to the global bias. Motivated by the GDPR's ``right to explanation'', research on explaining model predictions~\cite{ribeiro2016should,lundberg2017unified,lundberg2020local2global} has surged, but explaining prediction bias has received less attention~\cite{begley2020explainability,lundberg2020explaining}. In order to identify and explain the sources of bias and also the impact of affirmative/punitive actions to alleviate/deteriorate bias, it is important to understand \textit{which features contribute how much to the bias of a classifier} applied on a dataset. To this end, we follow a global feature-attribution approach to explain the sources of bias, where we relate the \emph{influences} of input features towards the resulting bias of the classifier. In this context, existing bias attributing methods~\cite{begley2020explainability,lundberg2020explaining} are variants of local function approximation~\cite{sliwinski2019axiomatic}, whereas bias is a global statistical property of a classifier. Thus, \textit{we aim to design a bias attribution method that is global by construction}. In addition, existing methods only attribute the individual influence of features on bias while neglecting the \textit{intersectionality} among features. Quantifying intersectionality allows us to explain bias induced by the higher-order interactions among features; hence accounting for intersectionality is important to understand bias as suggested by recent literature~\citep{buolamwini2018gender,wang2022towards}. 
In this paper, \textit{we aim to design a global bias attribution framework and a corresponding algorithm that can quantify both the individual and intersectional influences of features leading to granular and functional explanation of the sources of bias.}

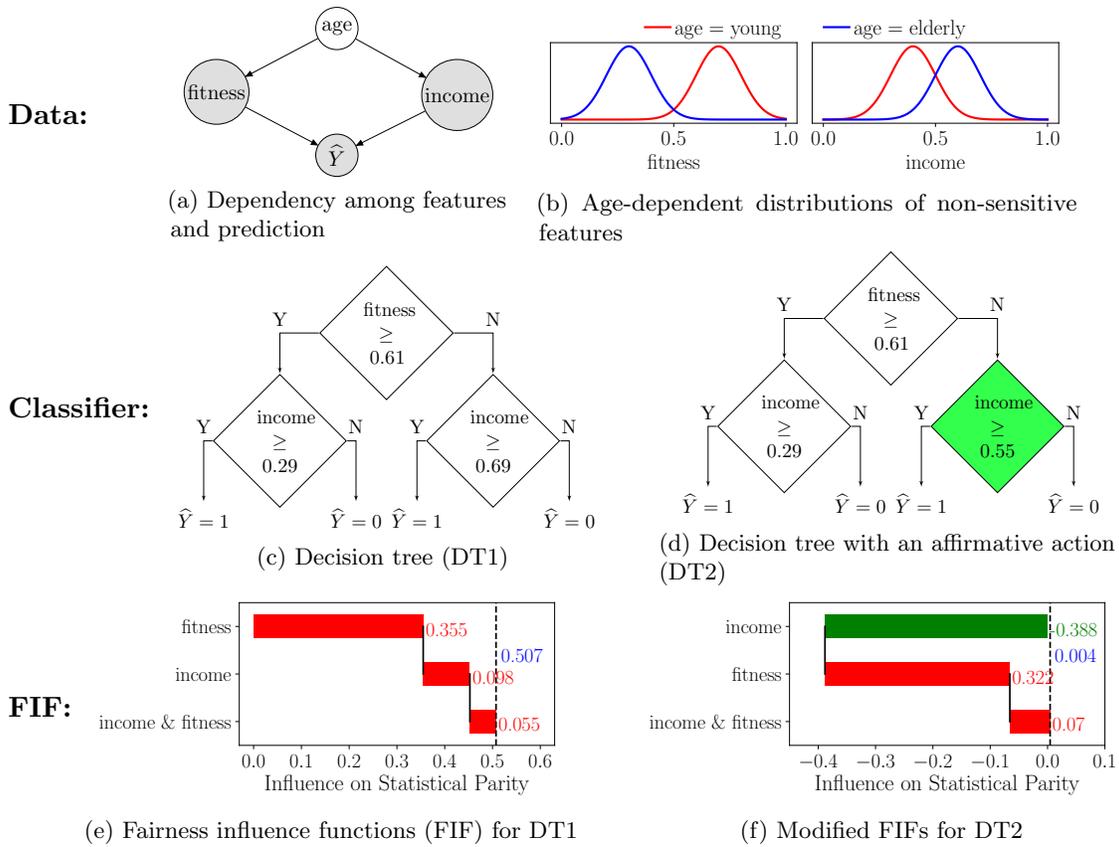
\begin{figure*}
	\begin{minipage}[t]{0.13\textwidth}			
		\scalebox{1}{	
		\begin{tikzpicture}[x=1cm,y=0.3cm]
		\node[] (a1) {\textbf{Data:}};			
		\end{tikzpicture}
		}
	\end{minipage}%
	\begin{minipage}{0.35\textwidth}
		\centering
		\subfloat[Dependency among features and prediction]{
			\scalebox{0.4}{	
				\begin{tikzpicture}[x=1.5cm,y=0.3cm]
				\node[latent,scale=2] (a1) {$\textrm{age}$} ; %
				\node[obs, scale=2, below=of a1, xshift=-2cm] (h) {$\textrm{fitness}$}; %
				\node[obs, scale=2, below=of a1, xshift=2cm] (i) {$\textrm{income}$}; %
				\node[obs, scale=2, below=of h, xshift=2cm] (p) {$\widehat{Y}$}; %
				\edge[] {a1} {h,i} ;
				\edge[] {h,i} {p} ;
				\end{tikzpicture}
			}	
			\label{fig:dag_age_income_fitness}}
	\end{minipage}%
	\begin{minipage}{0.4\textwidth}
		\centering
		\subfloat[Age-dependent distributions of non-sensitive features]{\includegraphics[scale=0.4]{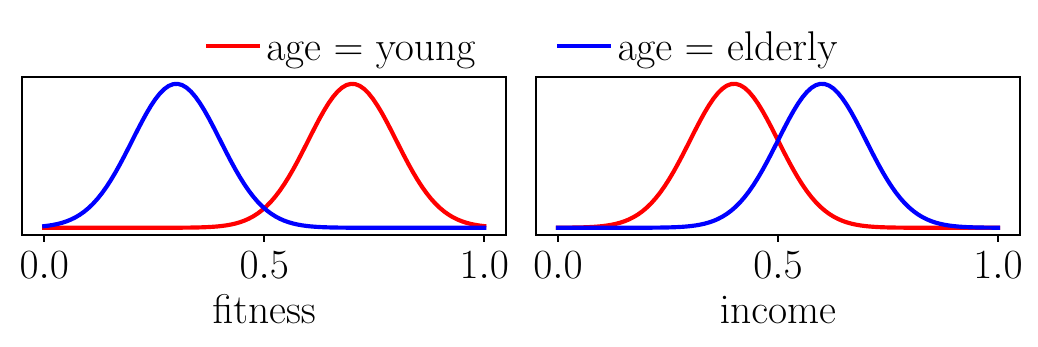}
		\label{fig:distribution_example}}
	\end{minipage}

	\begin{minipage}[t]{0.13\textwidth}
		\scalebox{1}{	
			\begin{tikzpicture}[x=1cm,y=0.3cm]
			\node[] (a1) {\textbf{Classifier:}};			
			\end{tikzpicture}
		}
	\end{minipage}%
	\begin{minipage}{0.43\textwidth}
		\centering
		\subfloat[Decision tree (DT$ 1 $)]{
			\scalebox{0.45}{	
				\begin{tikzpicture}[x=1cm,y=1.8cm]
				\node [box, scale=1.5]                                    (p)      {fitness $\geq 0.61$};
				\node [scale=1.5, box, below= of p, xshift=-2.1cm, yshift=1.2cm]    (a1)    {income\\ $\geq 0.29$};
				\node [scale=1.5, box, below= of p, xshift=2.1cm, yshift=1.2cm]     (a2)    {income $\geq 0.69$};
				\node [scale=1.5,below= of a1, xshift=-1.5cm, yshift=0.8cm]  (a11)    { $\widehat{Y}= 1$};
				\node [scale=1.5,below= of a1, xshift=1.5cm, yshift=0.8cm]   (a12)    { $\widehat{Y}=0 $};
				\node [scale=1.5,below= of a2, xshift=-1.5cm, yshift=0.8cm]  (a21)    { $\widehat{Y}= 1$};
				\node [scale=1.5,below= of a2, xshift=1.5cm, yshift=0.8cm]  (a22)    { $\widehat{Y}= 0$};
				\path [line] (p) -|         (a1) node [scale=1.5,midway, above]  {Y};
				\path [line] (p) -|         (a2) node [scale=1.5,midway, above]  {N};
				\path [line] (a1) -|       (a11) node [scale=1.5,midway, above]  {Y};
				\path [line] (a1) -|       (a12) node [scale=1.5,midway, above]  {N};
				\path [line] (a2) -|       (a21) node [scale=1.5,midway, above]  {Y};
				\path [line] (a2) -|       (a22) node [scale=1.5,midway, above]  {N};
				\end{tikzpicture}}
			\label{fig:dt_original}}
	\end{minipage}%
	\begin{minipage}{0.45\textwidth}
		\centering
		\subfloat[Decision tree with an affirmative action (DT$ 2 $)]{
			\scalebox{0.45}{	
				\begin{tikzpicture}[x=1cm,y=1.8cm]
				\node [box, scale=1.5]                                    (p)      {fitness $\geq 0.61$};
				\node [scale=1.5, box, below= of p, xshift=-2.1cm, yshift=1.2cm]    (a1)    {income\\ $\geq 0.29$};
				\node [scale=1.5, box, below= of p, xshift=2.1cm, yshift=1.2cm, fill=affirmative]     (a2)    {income $\geq 0.55$};
				\node [scale=1.5,below= of a1, xshift=-1.5cm, yshift=0.8cm]  (a11)    { $\widehat{Y}= 1$};
				\node [scale=1.5,below= of a1, xshift=1.5cm, yshift=0.8cm]   (a12)    { $\widehat{Y}=0 $};
				\node [scale=1.5,below= of a2, xshift=-1.5cm, yshift=0.8cm]  (a21)    { $\widehat{Y}= 1$};
				\node [scale=1.5,below= of a2, xshift=1.5cm, yshift=0.8cm]  (a22)    { $\widehat{Y}= 0$};
				\path [line] (p) -|         (a1) node [scale=1.5,midway, above]  {Y};
				\path [line] (p) -|         (a2) node [scale=1.5,midway, above]  {N};
				\path [line] (a1) -|       (a11) node [scale=1.5,midway, above]  {Y};
				\path [line] (a1) -|       (a12) node [scale=1.5,midway, above]  {N};
				\path [line] (a2) -|       (a21) node [scale=1.5,midway, above]  {Y};
				\path [line] (a2) -|       (a22) node [scale=1.5,midway, above]  {N};
				\end{tikzpicture}}	
			
			\label{fig:dt_affirmative}}
	\end{minipage}

	\begin{minipage}[t]{0.07\textwidth}
		\scalebox{1}{	
			\begin{tikzpicture}[x=1cm,y=0.3cm]
			\node[] (a1) {\textbf{FIF:}};			
			\end{tikzpicture}
		}
	\end{minipage}%
	\begin{minipage}{0.46\textwidth}
		\centering
		\subfloat[Fairness influence functions (FIF) for DT$ 1 $]{\includegraphics[scale=0.42]{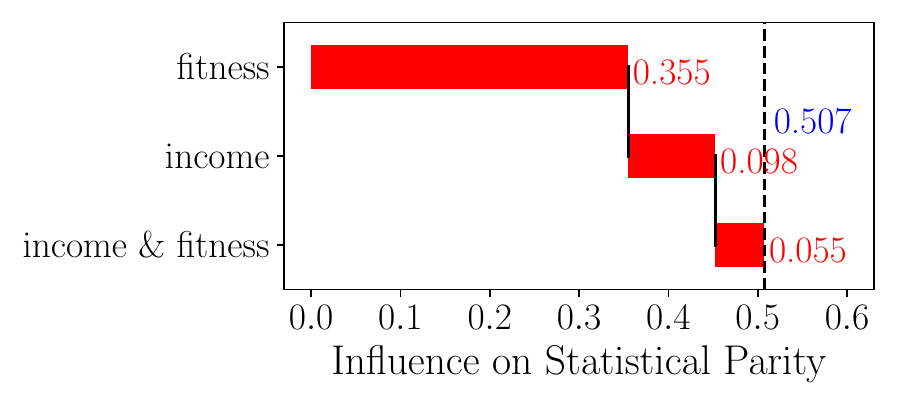}
		\label{fig:fif_original}}	
	\end{minipage}%
	\begin{minipage}{0.5\textwidth}
		\centering
		\subfloat[Modified FIFs for DT$ 2 $]{\includegraphics[scale=0.42]{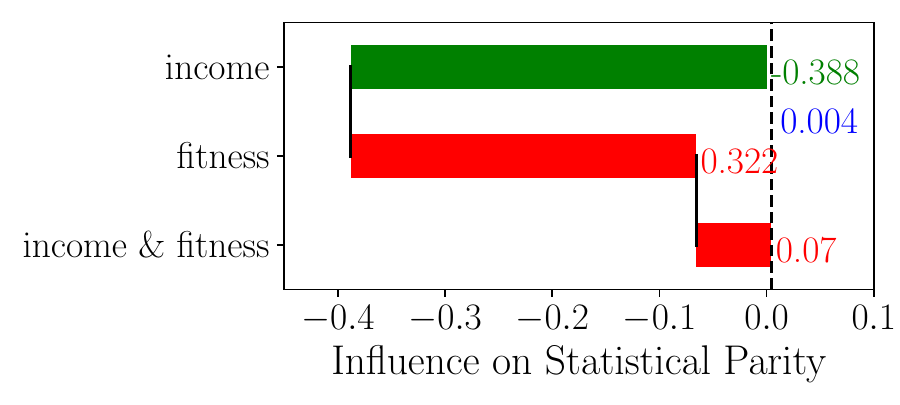}
		\label{fig:fif_affirmative}}
	\end{minipage}
	\caption{FIFs of input features to investigate the bias (statistical parity) of a decision tree predicting the eligibility for health insurance using age-dependent features `fitness' and `income'. An affirmative action reduces bias as corresponding FIFs reflect it.}
	\label{fig:fair_example_fif}
\end{figure*}

\paragraph{Contributions.}  Our contributions are three-fold.

\begin{enumerate}[leftmargin=*]
	
	\item \textbf{Formalism:} We propose to measure the contribution of individual and intersectional features towards the bias of a classifier operating on a dataset by estimating their \textbf{F}airness \textbf{I}nfluence \textbf{F}unctions (FIFs) (Section~\ref{sec:fifs}). Our method is based on transforming existing fairness metrics into the difference of the scaled conditional variances of classifier's prediction, which we then decompose using Global Sensitivity Analysis (GSA)\textemdash a standard technique recommended by regulators to assess numerical models~\citep{eu,usepa}. FIFs have several desirable properties (Theorem~\ref{thm:fif_property}), including the decomposability property~\cite{begley2020explainability,lundberg2020explaining}. This property states that the sum of FIFs of all individual and intersectional features equals the bias of the classifier. With this formulation of FIFs, we can identify which features have the greatest influence on bias by looking at the disparity in their scaled decomposed variance between sensitive groups.

	\item \textbf{Algorithmic:} We propose a new algorithm, called {\framework}, to estimate individual and intersectional FIFs of features given a dataset, a classifier, and a fairness metric. The algorithm is capable of working with any linear group fairness metric, including statistical parity, equalized odds, or predictive parity (Section~\ref{sec:fairxplainer}). Building on GSA~\citep{saltelli2008global} techniques, {\framework} solves a local regression problem~\cite{loader2006local} based on cubic splines~\cite{li2010global} to decompose the variance of the classifier's prediction among all the subsets of features.

	\item \textbf{Experimental:} We evaluate  {\framework}  on a variety of real-world datasets and machine learning classifiers to demonstrate its efficiency in estimating the individual and intersectional FIFs of features. Our results show that {\framework} has a higher accuracy in approximating bias using estimated FIFs compared to existing methods (Section~\ref{sec:experiments}). Our estimation of FIFs also shows a strong correlation with fairness interventions. Furthermore, {\framework} yields more granular explanation of the sources of bias by combining both individual and intersectional FIFs, and also detects patterns that existing fairness explainers cannot. Finally, {\framework} enables us to observe changes in FIFs as a result of different fairness enhancing algorithms~\cite{calmon2017optimized,hardt2016equality,kamiran2012decision,zemel2013learning,zhang2018mitigating,zhang2018fairness,zhang2019faht} and fairness reducing attacks~\cite{hua2021human,mehrabi2020exacerbating,solans2020poisoning}. This creates opportunities to further examine the impact of these algorithms.	
\end{enumerate}

We illustrate the usefulness of our contributions via an example scenario proposed in~\citep{ghosh2020justicia}.

\begin{example}
	\label{ex:motivating_example}  Following~\citep{ghosh2020justicia}, we consider a classifier that decides an individual's eligibility for health insurance based on non-sensitive features: fitness  and income. Fitness and income depend on a sensitive feature age $ \in $ \{young, elderly\} leading to two sensitive groups, as highlighted in (Figure~\ref{fig:dag_age_income_fitness}--\ref{fig:distribution_example}). 
	
	\textbf{Case study 1:} For each sensitive group, we generate $ 1000 $ samples of (income, fitness) and train a decision tree (DT$ 1 $), which predicts without explicitly using the sensitive feature (Figure~\ref{fig:dt_original}). %
	Using the $1000$ samples and corresponding predictions, we estimate the statistical parity of DT1 as $ \Pr[\widehat{Y} = 1  \mid  \text{age = young} ] - \Pr[\widehat{Y} = 1  \mid  \text{age = elderly}] = 0.695 - 0.165 = 0.53 $. Therefore, DT$ 1 $ is unfair towards the elderly group.

	Using the methods described in this paper, we examine the sources of unfairness in DT$ 1 $ by calculating the FIFs of each feature subset. Positive values represent a reduction in fairness due to increased statistical parity, while negative values indicate an improvement in fairness. The results, shown in the waterfall diagram in Figure~\ref{fig:fif_original}, indicate that fitness $(\mathrm{FIF} = 0.355)$, income $(\mathrm{FIF} = 0.098)$, and the combination of income and fitness $(\mathrm{FIF} = 0.055)$ contribute to higher statistical parity and bias. Fitness, being the root of DT$ 1 $, has a greater impact on statistical parity than income, which is at the second level of DT$ 1 $. Our method also reveals the joint effect of income and fitness on statistical parity, which prior methods do not account for (Section~\ref{sec:related_work}). The total of the FIFs of all features, $ 0.355 + 0.098 + 0.055 = 0.507 \approx 0.53 $, approximately matches the statistical parity of the classifier, providing a way to break down the bias among all feature subsets. Note that we discuss the estimation error of FIFs in Section~\ref{sec:fairxplainer}.

	\textbf{Case study 2:} 	
	To address the unfairness of DT$ 1 $ towards the elderly, we introduce DT$ 2 $ which applies an affirmative action. Specifically, for lower fitness, which is typical for the elderly group (Figure~\ref{fig:distribution_example}), we decrease the threshold on income from $ 0.69 $ to $ 0.55 $ ({\color{affirmative}green} node in Figure~\ref{fig:dt_affirmative}). This allows more elderly individuals to receive insurance as they tend to have higher income, and the lower threshold accommodates their eligibility. The statistical parity of DT$ 2 $ is calculated to be $ \Pr[\widehat{Y} = 1 \mid \text{age = young}] - \Pr[\widehat{Y} = 1 \mid \text{age = elderly}] = 0.707 - 0.706 = 0.001 $, which is negligible compared to the earlier statistical parity of $ 0.53 $ in DT$ 1 $. We estimate the FIFs of features, with $ - 0.388 $ for income, $ 0.322 $ for fitness, and $ 0.07 $ for both features combined. Hence, the negative influence of income confirms the affirmative action, and nullifies the disparity induced by fitness. Additionally, the sum of FIFs $ - 0.388 + 0.322 + 0.07 = 0.004 $ coincides with the statistical parity of DT$ 2 $. 
	
\end{example}

\section{Related Work}
\label{sec:related_work}
Recently, local explanation methods have been applied to black-box classifiers to explain sources of bias through feature-attribution~\citep{begley2020explainability,lundberg2020explaining} and causal path decomposition~\citep{pan2021explaining}. Our work uses the feature-attribution approach and highlights three limitations of existing methods: (i) a failure to estimate intersectional FIFs, (ii) inaccuracies in approximating bias based on FIFs, and (iii) less correlation of FIFs with fairness intervention. Elaborately, to explain group fairness\textemdash a global property of a classifier\textemdash existing local explanation approaches such as ~\cite{begley2020explainability,lundberg2020explaining} estimate FIFs based on a local black-box explainer SHAP~\cite{lundberg2017unified}. They apply a global aggregation (i.e. expectation) of all local explanations corresponding to all data points in a dataset. Such a global aggregation of local explanations is often empirically justified and does not approximate bias accurately (Section~\ref{sec:experiments}).  In addition, existing methods ignore the joint contribution of correlated features on bias. To address these limitations, \textit{we develop a formal framework to explain sources of group unfairness in a classifier and also a novel methodology to estimate FIFs. To the best of our knowledge, this is the first work to do both}.

Among other related works, \cite{benesse2021fairness} links GSA measures such as Sobol and Cram{\'e}r-von-Mises indices to different fairness metrics. While their approach relates the GSA of sensitive features on bias, we focus on applying GSA to all features to estimate FIFs. \textit{Their approach detects the presence or absence of bias, while we focus on decomposing bias as the sum of FIFs of all feature subsets}. In another line of work,~\cite{datta2016algorithmic} and~\cite{ghosh2022algorithmic} estimate feature-influence as the shifting of bias from its original value by randomly intervening features. Their work is different from the decomposability property of FIFs, where the sum of FIFs is equal to the bias. A separate line of work estimates the fairness influence of data points in a dataset~\cite{li2022achieving,wang2022understanding}, while \textit{our focus is on quantifying influences of input features}.

\section{Background: Fairness and Global Sensitivity Analysis}\label{sec:preliminaries}%
Before proceeding to the details of our contribution, we present the fundamentals of group fairness metrics as quantifiers of bias and global sensitivity analysis as a classical method of feature attribution.%
\subsection{Fairness in Machine Learning: Fairness Metrics}

We consider\footnote{{We represent sets/vectors by bold letters, and the corresponding distributions by calligraphic letters. We express random variables in uppercase, and an assignment of a random variable in lowercase.}} a dataset $ \mathbf{D} $ as a set of $n$ data points  $\{(\mathbf{z}^{(i)}, y^{(i)})\}_{i=1}^n$  generated from an underlying distribution $\mathcal{D}$. For the $ i^{\text{th}} $ data point, the feature vector $\mathbf{z}^{(i)} \triangleq (\mathbf{x}^{(i)}, \mathbf{a}^{(i)}) $ is a concatenation of non-sensitive features $ \mathbf{x}^{(i)} $ and sensitive features $ \mathbf{a}^{(i)} $. Each non-sensitive data point $\mathbf{x}^{(i)}$ consists of $\numnonsensitive$ features $[x^{(i)}_1, \dots, x^{(i)}_{\numnonsensitive}] \in \mathbb{R}^{\numnonsensitive} $. Each sensitive data point $\mathbf{a}^{(i)}$ consists of $\numsensitive$ categorical features $[a^{(i)}_1, \dots, a^{(i)}_{\numsensitive}] \in \mathbb{N}^{\numsensitive} $. Thus, the cardinality of feature vector $ \mathbf{z}^{(i)} $ is denoted by $ m $, formally $ |\mathbf{z}^{(i)}| \triangleq \numfeatures = \numnonsensitive + \numsensitive $. The binary class corresponding to $(\mathbf{x}^{(i)}, \mathbf{a}^{(i)})$ is $y^{(i)} \in \{0,1\}$. We refer to $y^{(i)}$ as the true class. We denote the random variables corresponding to $ (\mathbf{z}, \mathbf{x}, \mathbf{a},  y) $ with $ (\feature, \nonsensitive, \sensitive, Y) $.  Henceforth, we represent a binary classifier trained on dataset $\mathbf{D}$ as $\alg: (\nonsensitive, \sensitive) \rightarrow \widehat{Y} \in \{0,1\} $, where $\widehat{Y} $ is the predicted class. Given the setup, we discuss a fairness metric $ f: (\alg, \mathbf{D}) \rightarrow [0, 1] $ quantifying the bias in the prediction of a classifier on a dataset~\cite{feldman2015certifying,hardt2016equality,verma2018fairness}.

	\textit{Statistical Parity} ($ \mathsf{SP} $): Statistical parity belongs to the \textit{independence} measuring group fairness metrics that measures whether the prediction $ \widehat{Y} $ is statistically independent of the sensitive features $ \sensitive $.  The statistical parity of  a classifier is measured as $ f_{\mathsf{SP}}(\alg, \mathbf{D}) \triangleq \max_{\mathbf{a}}\Pr[\widehat{Y} =1 \mid \mathbf{A} = \mathbf{a}] - \min_{\mathbf{a}}\Pr[\widehat{Y} =1 \mid \mathbf{A} = \mathbf{a}] \in [0, 1] $, which is the difference between the maximum and minimum conditional probability of positive prediction the classifier for different sensitive groups. Lower value of $ f(\alg, \mathbf{D}) $ indicates higher fairness demonstrated by the classifier $\mathcal{M}$ on $ \mathbf{D} $. \textit{Henceforth, we deploy statistical parity as the measure of bias of a classifier w.r.t.\ a given dataset.}\footnote{For brevity, we define other  fairness metrics equalized odds and predictive parity, and corresponding FIF computation technique in the extended \href{https://arxiv.org/pdf/2206.00667.pdf}{paper}.}

\subsection{Global Sensitivity Analysis (GSA): Variance Decomposition}
Global sensitivity analysis is an active field of research that studies how the global uncertainty in the output of a function can be attributed to the different sources of uncertainties in the input variables~\citep{saltelli2008global}.
Sensitivity analysis is an essential component for quality assurance and impact assessment of models in EU~\citep{eu}, USA~\citep{usepa}, and research communities~\citep{saltelli2020five}.
\emph{Variance-based sensitivity analysis} is a form of global sensitivity analysis, where variance is considered as the measure of uncertainty~\citep{sobol1990sensitivity,sobol2001global}. To illustrate, let us consider a real-valued function $  \function(\feature) $, where $ \feature $ is a vector of $ \numfeatures $ input variables $ \{Z_1, \dots, Z_\numfeatures\} $.
Now, we decompose $ \function(\feature) $ among the subsets of inputs, such that:
\begin{align}
	\function(\feature) &= \function_0 + \sum_{i=1}^{\numfeatures} \function_{\{i\}}(Z_i) +  \sum_{i < j}^{\numfeatures} \function_{\{i,j\}}(Z_i, Z_j)  + \cdots\notag \\&\quad\quad\quad\quad+ \function_{\{1, 2, \dots, \numfeatures\}} (Z_1, Z_2, \dots, Z_{\numfeatures})\\
	&= \function_0 +  \sum_{\mathbf{S} \subseteq [\numfeatures]} \function_{\mathbf{S}}(\feature_{\mathbf{S}})\label{eq:functional_decomposition_set_notation}
\end{align}
The standard condition of this decomposition is the orthogonality of each term in the right-hand side of Eq.~\eqref{eq:functional_decomposition_set_notation}~\citep{sobol1990sensitivity}. In this decomposition, $ \function_0 $ is a constant, $ \function_{\{i\}} $ is a function of $ Z_i $, $ \function_{\{i,j\}} $ is a function of $ Z_i $ and $ Z_j $, and so on. Adhering to the set-based notations, we denote by $ g_{\mathbf{S}} $ a function of a non-empty subset of variables $ \feature_{\mathbf{S}}  \triangleq \{Z_i \mid i \in \mathbf{S}\} \subseteq \feature $, where $ \mathbf{S} = \{S_i \mid 1 \le |S_i| \le \numfeatures\} \subseteq [\numfeatures] $ is a non-empty subset of indices with $ [\numfeatures] \triangleq \{1,2,\dots, \numfeatures\}  $.  Here, $|\mathbf{S}|=1$ quantifies an individual variable's effect while $|\mathbf{S}|>1$ quantifies the higher-order intersectional effect of variables.
Considering $ \function $ as square integrable, we obtain the decomposition of the variance of $ \function(\feature) $ expressed as the sum of variances of $\function_{\mathbf{S}}$'s~\citep{sobol1990sensitivity}.
\begin{align}\label{eq:variance_decomposition_set_notation}
	\mathsf{Var}[g(\feature)] &= \sum_{i=1}^{\numfeatures}V_{\{i\}} +  \sum_{i<j}^{\numfeatures} V_{\{i,j\}}  + \dots +  V_{\{1, 2\dots, \numfeatures\}}= \sum_{\mathbf{S} \subseteq [\numfeatures]} V_{\mathbf{S}} 
\end{align}
where $ V_{\{i\}} $ is the variance of $ \function_{\{i\}} $, $ V_{\{i,j\}} $ is the variance of $ \function_{\{i,j\}} $ and so on. Formally,  
\begin{equation}
\label{eq:decomposed_variance}
V_{\mathbf{S}} \triangleq \mathsf{Var} _{\feature_{\mathbf{S}}}\left[\mathbb{E}_{{\feature}\setminus\feature_{\mathbf{S}}}[g(\feature)\mid \feature_{\mathbf{S}}]\right]- \sum_{\mathbf{S}' \subset \mathbf{S}}{V} _{\mathbf{S}'}.
\end{equation} 
Here, $\mathbf{S}'$ denotes all the non-empty and ordered proper subsets of $\mathbf{S}$. Thus, $ V_{\mathbf{S}} $ is the variance w.r.t.\ $ \feature_{\mathbf{S}} $ by subtracting the variances of all non-empty proper subsets of $ \feature_{\mathbf{S}} $. As a result, Eq.~\eqref{eq:variance_decomposition_set_notation} demonstrates how the variance of $\function(\feature) $ can be decomposed into terms attributable to each input feature, as well as the intersectional effects among them\footnote{In spirit, variance decomposition is an extension of functional ANOVA introduced by~\citep{hoeffding1992class} and studied by~\cite{efron1981jackknife} and~\cite{sobol1969multidimensional}.}. Conversely, together all terms sum to the total variance of the model output. 
\textit{In the next section, we reduce the problem of computing FIFs of subsets of features into a variance decomposition problem.}

\section{Fairness Influence Functions: Formulation and Properties}\label{sec:fifs}
We formalize Fairness Influence Functions (FIF) as a quantifier of the contribution of a subset of features to the resultant bias of a classifier applied on a dataset.  In GSA, we observe that the variance of the output of a function can be attributed to the corresponding subset of input variables through variance decomposition (Eq.~\eqref{eq:variance_decomposition_set_notation}). To leverage the power of GSA in fairness in machine learning, we first express the existing fairness metrics in terms of the variance of classifier's prediction in Section~\ref{sec:metric_as_variance}. This allows us to formulate FIF by leveraging variance decomposition (Section~\ref{sec:fif_formulation}).

\subsection{Fairness Metrics as the Variance of Prediction}
\label{sec:metric_as_variance}

First, we observe that the random variable central to computing statistical parity is $ \mathds{1}[\widehat{Y} = 1 \mid \sensitive = \mathbf{a}] $.
We refer to this indicator function as \textit{Conditional Positive Prediction (CPP)} of a classifier. Now, we express statistical parity as a functional of the probability of CPPs for different sensitive groups~\cite{ghosh2020justicia,benesse2021fairness}. For brevity, we defer similar formulations for the other fairness metrics, i.e. equalized odds and predictive parity, to the extended \href{https://arxiv.org/pdf/2206.00667.pdf}{paper}. %

Our key idea for computing FIFs of features is to represent fairness metrics using the variance of CPPs. Formally, we express statistical parity using the variance of CPPs in Lemma~\ref{lm:sp_var_relation}.

\begin{lemma}[Statistical Parity as Difference of Variances of CPPs]
	\label{lm:sp_var_relation}
	Let $ \mathbf{a}_{\max} = \argmax_{\mathbf{a}} \Pr[\widehat{Y} = 1 \mid  \sensitive = \mathbf{a}] $ and $ \mathbf{a}_{\min} = \argmin_{\mathbf{a}} \Pr[\widehat{Y} = 1 \mid \sensitive = \mathbf{a}] $ be the most and the least favored sensitive groups, respectively. The statistical parity of a binary\footnote{For a multi-class classifier, statistical parity of the target class $ y \in \mathbb{N} $ is $ \frac{\mathsf{Var}[\widehat{Y} = y \mid \sensitive = \mathbf{a}_{\max}]}{1 - \Pr[\widehat{Y} = y \mid  \sensitive = \mathbf{a}_{\max}]} - \frac{\mathsf{Var}[\widehat{Y} = y \mid \sensitive = \mathbf{a}_{\min}] }{1 - \Pr[\widehat{Y} = y \mid  \sensitive = \mathbf{a}_{\min}]}. $} classifier is the difference in the scaled variance of CPPs. Formally, if $\Pr[\widehat{Y} = 0 \mid  \sensitive = \mathbf{a}_{\max}] \neq 0$ and $\Pr[\widehat{Y} = 0 \mid  \sensitive = \mathbf{a}_{\min}] \neq 0$,
	\[
	 f_{\mathsf{SP}}(\alg, \mathbf{D}) = \frac{\mathsf{Var}[\widehat{Y} = 1\mid\sensitive = \mathbf{a}_{\max}]}{\Pr[\widehat{Y} = 0 \mid  \sensitive = \mathbf{a}_{\max}]} - \frac{\mathsf{Var}[\widehat{Y} = 1\mid\sensitive = \mathbf{a}_{\min}] }{\Pr[\widehat{Y} = 0 \mid  \sensitive = \mathbf{a}_{\min}]}.
	\]
\end{lemma}
\begin{proof}
	For a sensitive group $ \mathbf{a} $, CPP  is a Bernoulli random variable, where probability\footnote{For any binary event $ E $, expectation and probability are identical, $ \mathbb{E}[\mathds{1}(E)] = \Pr[E = 1] $.} $ p_{\mathbf{a}}  = \Pr[\widehat{Y} = 1 \mid \sensitive = \mathbf{a}] $ and variance $ V_{\mathbf{a}} = \mathsf{Var}[\widehat{Y} = 1 \mid \sensitive = \mathbf{a}] = p_{\mathbf{a}} (1 - p_{\mathbf{a}}) $. Thus, for sensitive groups $ \mathbf{a}_{\max} $ and $ \mathbf{a}_{\min} $, the statistical parity of the classifier is $ p_{\mathbf{a}_{\max}}  - p_{\mathbf{a}_{\min}} =  \frac{V_{\mathbf{a}_{\max}}}{1 - p_{\mathbf{a}_{\max}}}  - \frac{V_{\mathbf{a}_{\min}}}{1 - p_{\mathbf{a}_{\min}}} $. Replacing $ 1 - p_\mathbf{a} = \Pr[\widehat{Y} = 0 \mid \sensitive = \mathbf{a}] $  proves the lemma.
\end{proof}

\begin{repexample}{ex:motivating_example}[Revisited]
	From Figure~\ref{fig:dt_original}, the probability of CPPs of the decision tree for the most and least favored groups are $  \Pr[\widehat{Y} = 1 \mid \text{age = young}] = 0.695 $ and $ \Pr[\widehat{Y} = 1 \mid \text{age = elderly}] = 0.165 $, respectively. Thus, the statistical parity is $ 0.695 - 0.165 =  0.53 $. Next, we compute variance of CPPs as $  \mathsf{Var}[\widehat{Y} = 1\mid \text{age = young}] = 0.212 $ and $  \mathsf{Var}[\widehat{Y} = 1\mid \text{age = elderly}] = 0.138 $. Thus, following Lemma~\ref{lm:sp_var_relation}, we compute the difference in the scaled  variance of CPPs as $ \frac{0.212}{1 - 0.695} - \frac{0.138}{1 - 0.165} =  0.529 \approx 0.53 $, which coincides with the statistical parity reported earlier.
\end{repexample}

\subsection{Formulation of FIF}
\label{sec:fif_formulation}
We are given a binary classifier  $\alg $, a dataset $ \mathbf{D} $, and a  fairness metric $ f(\alg, \mathbf{D}) \in [0, 1] $. Our objective is to compute the influences of  features on $ f $. Particularly, we compute the influence of each subset of  features $ \feature_{\mathbf{S}} $, where $ \mathbf{S} = \{S_i \mid 1 \le |S_i| \le \numfeatures\} \subseteq [\numfeatures]  $ is a non-empty subset of indices of $ \feature $.

\begin{definition} \textbf{Fairness Influence Function (FIF)}\footnote{For equalized odds, a precise definition of FIF is $ w_{\mathbf{S}}: (\alg, Y, \feature_{\mathbf{S}}) \rightarrow \mathbb{R} $ taking the ground class $ Y $ as an additional input. For predictive parity, FIF is defined as $ w_{\mathbf{S}}: (\alg, \widehat{Y}, \feature_{\mathbf{S}}) \rightarrow \mathbb{R} $ taking the predicted class $ \widehat{Y} $ as an additional input.}, denoted by $ w_{\mathbf{S}}: (\alg, \feature_{\mathbf{S}}) \rightarrow \mathbb{R} $, measures the quantitative contribution of features $ \feature_{\mathbf{S}} \subseteq \feature $ on the incurred bias $ f(\alg, \mathbf{D}) $. Leveraging the variance-difference representation of $ f(\alg, \mathbf{D}) $ (Lemma~\ref{lm:sp_var_relation}) and variance decomposition (Eq.~\eqref{eq:variance_decomposition_set_notation}), we particularly define $ w_{\mathbf{S}} $ as
\begin{equation}\label{eq:fif_decompose}
	w_{\mathbf{S}}  \triangleq \frac{V_{\mathbf{a}_{\max}, \mathbf{S}}}{ \Pr[\widehat{Y} = 0 \mid  \sensitive = \mathbf{a}_{\max}]} - \frac{V_{\mathbf{a}_{\min}, \mathbf{S}} }{\Pr[\widehat{Y} = 0 \mid  \sensitive = \mathbf{a}_{\min}]}.
\end{equation}
Here, given a sensitive group $ \mathbf{a} $, $ V_{\mathbf{a}, \mathbf{S}} \triangleq  \mathsf{Var} _{\feature_{\mathbf{S}}}[\mathbb{E}_{\feature\setminus\feature_{\mathbf{S}}}[\widehat{Y} = 1\mid \sensitive = \mathbf{a}, \feature_{\mathbf{S}}]]- \sum_{\mathbf{S}' \subset \mathbf{S}}{V} _{\mathbf{a}, \mathbf{S}'} $ is the contribution (decomposed variance) of features $ \feature_{\mathbf{S}} $ in $ \mathsf{Var}[\widehat{Y} = 1\mid\sensitive = \mathbf{a}] $.
\end{definition}	

Informally, the FIF $ w_{\mathbf{S}} $ of $ \feature_{\mathbf{S}} $ is the difference in the scaled decomposed variance of CPPs between sensitive groups $ \mathbf{a}_{\max} $ and $ \mathbf{a}_{\min} $ as induced by  $ \feature_{\mathbf{S}} $. Thus, FIF of features is non-zero when the scaled decomposed variance-difference of CPPs is non-zero for those features, and vice versa. We refer to $ w_{\mathbf{S}} $ as an \emph{individual influence} when $ |\mathbf{S}| = 1 $, and as an \emph{intersectional influence} when $ |\mathbf{S}| > 1 $. Being able to naturally quantify the higher-order influences allow FIFs to explain the sources of bias in a more fine-grained manner. We experimentally validate this in Section~\ref{sec:experiments}.

\paragraph{Properties of FIF} FIF as defined in Eq.~\eqref{eq:fif_decompose} yields interesting properties, such as decomposability\footnote{Decomposability property is also known as the efficiency property in the context of Shapley values~\cite{roth1988shapley}.}, symmetry, and null properties, which we formally state in Theorem~\ref{thm:fif_property}.

\begin{theorem}[Properties of FIF]
	\label{thm:fif_property}
	Let $ f(\alg, \mathbf{D}) $ be the bias/unfairness of the classifier $ \alg $ on dataset $ \mathbf{D} $ according to linear group fairness metrics such as statistical parity. Let $ w_{\mathbf{S}}  $ be the FIF of a subset of  features $ \feature_{\mathbf{S}} $ as defined in Eq.~\eqref{eq:fif_decompose}. 
	\begin{enumerate}
		\item[(a)] \textit{The decomposability property} of FIF states that the sum of FIFs of all subset of  features is equal to the bias of the classifier. 
		\begin{align}\label{eq:constraint_framework}
		\sum_{\mathbf{S} \subseteq [\numfeatures] } w_{\mathbf{S}} = f(\alg, \mathbf{D})
		\end{align}
		\item[(b)] \textit{The symmetry property} states that  two features $ Z_i $ and $ Z_j $ are equivalent based on FIF if the sum of corresponding individual influences and the intersectional influences with all other features are the same. Mathematically,
		\begin{align}
		\sum_{\mathbf{S}'' \subseteq [m]\setminus\{i,j\}}w_{\mathbf{S}''\cup \{i\}} = \sum_{\mathbf{S}'' \subseteq [m]\setminus\{i,j\}}w_{\mathbf{S}''\cup \{j\}}  
		\end{align}
		if $\sum_{\mathbf{S}' \subseteq \mathbf{S}\cup \{i\}} w_{\mathbf{S}'} = 		\sum_{\mathbf{S}' \subseteq \mathbf{S} \cup \{j\}} w_{\mathbf{S}'}$ for every non-empty subset $ \mathbf{S} $ of $ [\numfeatures] $ containing neither $ i $ nor $ j $. 
 		\item[(c)] \textit{The null property} of FIF states that feature $ Z_i $ s a dummy or neutral feature if sum of its individual influence and the intersectional influences with all other features is zero. Mathematically,
	\begin{align}
	 \sum_{\mathbf{S}'' \subseteq [m]\setminus \{i\}}w_{\mathbf{S}''\cup \{i\}} = 0 	 
	\end{align}
	if	$\sum_{\mathbf{S}' \subseteq \mathbf{S} \cup \{i\}} w_{\mathbf{S}'} = 		\sum_{\mathbf{S}' \subseteq \mathbf{S}} w_{\mathbf{S}'}$ for every non-empty subset $ \mathbf{S} $ of $ [\numfeatures] $ that does not contain $ i $.
\end{enumerate}
\end{theorem}
We emphasize that the decomposability property proposed here is global, i.e. it holds for a whole dataset, but the one used for Shapley-value based explanations~(ref. Def. 1, \citep{lundberg2017unified}), or any local explanation method~\cite{han2022explanation}, is specific to a given data point~\cite{sliwinski2019axiomatic}. Thus, they are fundamentally different.

The symmetry and null properties of FIFs also distinguish the FIF quantification proposed in Eq.~\eqref{eq:fif_decompose} from the attribution methods considering only individual features, like SHAP. For example, a feature $i$ has zero impact on bias if not only its individual influence but the sum of influences of all the subsets of features that include $i$ is zero. Thus, the symmetry and null properties stated here are by default global and intersectional, and these two aspects are absent in the existing bias explainers.

\paragraph{Bias Amplifying and Eliminating Features.} The sign of $ w_{\mathbf{S}} $ indicates whether features $ \feature_{\mathbf{S}} $ amplify the bias of the classifier or eliminate it. When the scaled decomposed variance of CPPs w.r.t.\ features $ \feature_{\mathbf{S}} $ for the sensitive group $ \mathbf{a}_{\max} $ is higher than the group $ \mathbf{a}_{\min} $, $ w_{\mathbf{S}} > 0 $. As such, $ \feature_{\mathbf{S}} $ increase bias. Conversely, when $ w_{\mathbf{S}} < 0 $, $ \feature_{\mathbf{S}} $ eliminates bias, and improves fairness. Finally, features $ \feature_{\mathbf{S}} $ are neutral in bias when $ w_{\mathbf{S}} = 0 $. 

Now, we present the following two propositions that relate the sign of $ w_{\mathbf{S}} $ with the decomposed variance of CPPs.

\begin{proposition}\label{prop:neg_fif}
	When $ w_{\mathbf{S}} < 0 $, i.e. features $ \feature_{\mathbf{S}} $ decrease bias, the decomposed variance of CPPs w.r.t.\ $ \feature_{\mathbf{S}} $ follows $ V_{\mathbf{a}_{\max}, \mathbf{S}} < V_{\mathbf{a}_{\min}, \mathbf{S}}  $.
\end{proposition}
Proposition~\ref{prop:neg_fif} implies that if a subset of features $ \feature_{\mathbf{S}} $ is bias-eliminating, the conditional variance in positive outcomes induced by $ \feature_{\mathbf{S}} $ is smaller for the most favoured group than that of the least favoured group.
\begin{proposition}\label{prop:pos_fif}
{If the decomposed variance of CPPs w.r.t.\ $ \feature_{\mathbf{S}} $ satisfies $ V_{\mathbf{a}_{\max}, \mathbf{S}} > {V_{\mathbf{a}_{\min}, \mathbf{S}}} $, the corresponding FIF $ w_{\mathbf{S}} > 0  $, i.e. features $ \feature_{\mathbf{S}} $ increase bias.}
\end{proposition}
Proposition~\ref{prop:pos_fif} implies that if the conditional variance in positive outcomes induced by $ \feature_{\mathbf{S}} $ is larger for the most favoured group than that of the least favoured group, then this subset of features $ \feature_{\mathbf{S}} $ is bias-inducing.

\paragraph{Special Cases.} Since our FIF formulation is based on the variance of prediction, for (degenerate) cases when the conditional prediction of the classifier is always positive or always negative for any sensitive group, the variance of prediction becomes zero. This observation results in following two propositions.

\begin{proposition}[Perfectly Unbiased Classifiers]
	When $ \Pr[\widehat{Y} = 1 \mid \sensitive = \mathbf{a}_{\max} ] = \Pr[\widehat{Y} = 1 \mid \sensitive = \mathbf{a}_{\min} ] $ and both conditional probabilities are either $ 0 $ or $ 1 $, the FIF $w_{\mathbf{S}}  = 0 $ for all subsets of features $ \feature_{\mathbf{S}} $.
\end{proposition}
When $ \Pr[\widehat{Y} = 1 \mid \sensitive = \mathbf{a}_{\max} ] = \Pr[\widehat{Y} = 1 \mid \sensitive = \mathbf{a}_{\min} ] $ for a classifier, it means that the classifier equally yields positive predictions for each of the sensitive groups. Thus, there is no bias (in terms of statistical parity) in the classifier outcome. In that case, our formulation of FIF yields $w_{\mathbf{S}}  = 0 $ for all the subsets of features, and leads to a degenerate conclusion that all subsets of features are neutral or zero-bias inducing.
\begin{proposition}[Perfectly Biased Classifier]
	{When statistical parity is $ 1 $, i.e. $ \Pr[\widehat{Y} = 1 \mid \sensitive = \mathbf{a}_{\max} ] = 1 $ and $ \Pr[\widehat{Y} = 1 \mid \sensitive = \mathbf{a}_{\min} ] = 0 $, the sensitive features $ \sensitive $ are solely responsible for bias. In this case, the FIF $ w_{\mathbf{S}} = 0 $ for all features.}
\end{proposition}
When a classifier always yields positive predictions for the most favoured group and only negative predictions for the least favoured group, we obtain $ \Pr[\widehat{Y} = 1 \mid \sensitive = \mathbf{a}_{\max} ] = 1 $ and $ \Pr[\widehat{Y} = 1 \mid \sensitive = \mathbf{a}_{\min} ] = 0 $. This is a perfectly biased classifier, which can be expressed by the binary rule $\widehat{Y} = 1$ if $\sensitive = \mathbf{a}_{\max}$ and $0$ otherwise. As the discrimination is only due to the sensitive features, our FIF formulation cannot attribute any bias to the any other subset of  features, which leads to $ w_{\mathbf{S}} = 0 $ for all $\mathbf{S}$.

\textit{Expressing FIF in terms of the variance decomposition over a subset of features allows us to import and extend well-studied techniques of GSA to perform FIF estimation}, which we elaborate on Section~\ref{sec:fairxplainer}.

\begin{algorithm*}
	\caption{{\framework}}\label{algo:framework}
	\begin{flushleft}
		\hspace*{\algorithmicindent}\textbf{Input:} Classifier $ \alg: (\nonsensitive, \sensitive) \rightarrow \widehat{Y}$, Dataset $\mathbf{D} = \{(\mathbf{z}^{(i)}, y^{(i)})\}_{i=1}^n$, Fairness metric $f(\alg, \mathbf{D}) \in [0, 1] $, Maximum order of intersectional influence $\lambda $\\
		\hspace*{\algorithmicindent}\textbf{Output:} FIF $ w_{\mathbf{S}}$ for the subsets of features $\{\feature_{\mathbf{S}}\}$
	\end{flushleft}
	
	\begin{algorithmic}[1]
		\State $ \mathbf{a}_{\max} = \argmax_{\mathbf{a}} \Pr[\widehat{Y} = 1 | \sensitive = \mathbf{a}], \mathbf{a}_{\min} = \argmin_{\mathbf{a}} \Pr[\widehat{Y} = 1 | \sensitive = \mathbf{a}], \numfeatures \gets    |\feature|, \feature \equiv (\nonsensitive, \sensitive) $
		\label{algo_line:fif_computation_start}
		
		\For{$ \mathbf{a} \in \{\mathbf{a}_{\max}, \mathbf{a}_{\min}\} $} \Comment{Enumerate for specific sensitive groups}
		\State $ \function_{\mathbf{a}, \mathbf{S}},\function_{\mathbf{a}, 0} \gets \textsc{LocalRegression}(\alg(\nonsensitive, \sensitive = \mathbf{a}), \{\mathbf{z}^{(i)}\}_{i=1}^n, \lambda, m) $
		
		\State $ V_{\mathbf{a}, \mathbf{S}} \gets \textsc{Covariance}(\function_{\mathbf{a}},  \{\mathbf{z}^{(i)}\}_{i=1}^n, \function_{\mathbf{a}, \mathbf{S}},\function_{\mathbf{a}, 0}) $
		\EndFor
		\State Compute $ w_{\mathbf{S}}$ using  $V_{\mathbf{a}_{\max},\mathbf{S}}$ and $V_{\mathbf{a}_{\min},\mathbf{S}}$ as in Equation~\eqref{eq:fif_decompose}
		\label{algo_line:fif_computation_end}
		
		\Statex
		\Function{LocalRegression}{$ g_\mathbf{a}, \{\mathbf{z}^{(i)}\}_{i=1}^n, \lambda, m $}
		\label{algo_line:local_regression_start}
		\State \textbf{Initialize}: $ \function_{\mathbf{a}, 0} \leftarrow \textsc{Mean}(\{g(\mathbf{z}^{(i)})\}_{i=1, \mathbf{a}^{(i)} = \mathbf{a}}^n), \widehat{\function}_{\mathbf{a}, \mathbf{S}} \leftarrow 0, \forall \mathbf{S} \in [m], \mathbf{S} \ne \emptyset, |\mathbf{S}|\le \lambda$ \label{alg_line:initialization}
		\While{each $ \widehat{\function}_{\mathbf{a},\mathbf{S}} $ does not converge}
		\For{each $\mathbf{S}$}
		\State $ \widehat{\function}_{\mathbf{a}, \mathbf{S}} \leftarrow \textsc{Smooth}\big(\{\function_{\mathbf{a}}(\mathbf{z}^{(i)}) - \function_{\mathbf{a}, 0} - \sum_{\mathbf{S} \ne \mathbf{S}'} \widehat{\function}_{\mathbf{a}, \mathbf{S}'}(\mathbf{z}_{\mathbf{S}}^{(i)})\}_{i=1, \mathbf{a}^{(i)} = \mathbf{a}}^n\big)
		$\label{alg_line:backfitting_step} \Comment{Backfitting}
		\State $\widehat{\function}_{\mathbf{a}, \mathbf{S}}  \leftarrow \widehat{\function}_{\mathbf{a}, \mathbf{S}} - \textsc{Mean}\big(\{\widehat{\function}_{\mathbf{a}, \mathbf{S}}(\mathbf{z}_{\mathbf{S}}^{(i)})\}_{i=1, \mathbf{a}^{(i)} = \mathbf{a}}^n\big)$ \label{alg_line:mean_centered} \Comment{Mean centering}
		\EndFor
		\EndWhile
		\State \Return $ \function_{\mathbf{a}, \mathbf{S}},\function_{\mathbf{a}, 0} $
		\label{algo_line:local_regression_end} 
		\EndFunction

		\Function{Covariance}{$\function_{\mathbf{a}},  \{\mathbf{z}^{(i)}\}_{i=1}^n, \function_{\mathbf{a}, \mathbf{S}},\function_{\mathbf{a}, 0}$}
		\label{algo_line:covariance_computation_start}
		\State \Return $ \sum_{i=1, \mathbf{a}^{(i)} = \mathbf{a}}^n \function_{\mathbf{a}, \mathbf{S}}(\mathbf{z}_{\mathbf{S}}^{(i)})(g_\mathbf{a}(\mathbf{z}^{(i)}) - \function_{\mathbf{a}, 0}) $
		\label{algo_line:covariance_computation_end}
		\EndFunction
	\end{algorithmic}
\end{algorithm*}

\section{\framework: An Algorithm to Estimate Fairness Influence Functions}\label{sec:fairxplainer}

We propose an algorithm, {\framework}, that leverages the variance decomposition of CPPs to estimate the FIFs of all subsets of features. {\framework} has two algorithmic blocks: (i) local regression to decompose the classifier into component functions taking distinct subsets of features as input and (ii) computing the variance (or covariance) of each component function. We describe the schematic of \framework{} in Algorithm~\ref{algo:framework}.
\setlength{\textfloatsep}{12pt}%

\paragraph{A Set-additive Representation of the Classifier.} To apply variance decomposition (Eq.~\eqref{eq:variance_decomposition_set_notation}), we learn a set-additive representation of the classifier (Eq.~\eqref{eq:functional_decomposition_set_notation}) with input $ \feature \equiv (\nonsensitive, \sensitive) $. Let us denote the classifier $ \alg $ conditioned on a sensitive group $ \mathbf{a} $ as $ \function_{\mathbf{a}}(\feature) \triangleq \alg(\nonsensitive, \sensitive = \mathbf{a}) $. We express $ \function_{\mathbf{a}} $ as a set-additive model:
\begin{align}
\label{eq:set_additive_classifier}
\function_{\mathbf{a}}(\feature) = \function_{\mathbf{a}, 0} +  \sum_{\mathbf{S} \subseteq [\numfeatures] , |\mathbf{S}| \le \lambda} \function_{\mathbf{a},\mathbf{S}}(\feature_{\mathbf{S}}) + \delta
\end{align}

Here, $ \function_{\mathbf{a}, 0} $ is a constant, $ \function_{\mathbf{a},\mathbf{S}} $ is a \emph{component function} of $ \function_{\mathbf{a}} $ taking  a non-empty subset of features $ \feature_{\mathbf{S}} $ as input, and $ \delta $ is the approximation error. For computational tractability, we consider only components of \emph{maximum order} $ \lambda $, denoting the maximum order of intersectionality. {\framework} deploys backfitting algorithm for learning component functions in Eq.~\eqref{eq:set_additive_classifier}, as discussed in the following.

\paragraph{Local Regression with Backfitting.} We perform local regression with backfitting algorithm to learn the component functions up to a given order $ \lambda $ (Line~\ref{algo_line:local_regression_start}--\ref{algo_line:local_regression_end}). Backfitting algorithm is an iterative algorithm, where in each iteration one component function, say $ \function_{\mathbf{a}, \mathbf{S}} $, is learned while keeping other component functions fixed. Specifically, $ \function_{\mathbf{a}, \mathbf{S}} $ is learned as a smoothed function of $ g $ and rest of the components $ \function_{\mathbf{a}, \mathbf{S}'} $, where $ \mathbf{S}' \ne \mathbf{S} $ is a non-empty subset of $ [\numfeatures] $. To keep every component function mean centered, backfitting requires to impose two constraints: (i) $ \function_{\mathbf{a}, 0} = \textsc{Mean}\big(\{g(\mathbf{z}^{(i)})\}_{i=1, \mathbf{a}^{(i)} = \mathbf{a}}^n\big) $ (Line~\ref{alg_line:initialization}), which is the mean of $ \function_{\mathbf{a}} $ evaluated on data points belonging to the sensitive group $ \mathbf{a} $;  and (ii) $ \sum_{i=1, \mathbf{a}^{(i)} = \mathbf{a}}^n\function_{\mathbf{a}, \mathbf{S}}(\mathbf{z}_{\mathbf{S}}^{(i)}) = 0$ (Line~\ref{alg_line:mean_centered}), where $ \mathbf{z}_{\mathbf{S}}^{(i)} $ is the subset of feature values associated with feature indices $ \mathbf{S} $ for the $ i $-th data point $  \mathbf{z}^{(i)} $. These constraints assign the expectation of $ \function_{\mathbf{a}} $ on the constant term $ \function_{\mathbf{a}, 0} $ and the variance of $ \function_{\mathbf{a}} $ to the component functions.

While performing local regression, backfitting uses a smoothing operator~\citep{loader2012smoothing} over the set of data points (Line~\ref{alg_line:backfitting_step}). A smoothing operator, referred as $ \textsc{Smooth} $, allows us to learn a global representation of a component function by smoothly interpolating $\tau + 2$ local points obtained by local regression~\citep{loader2012smoothing}. In this paper, we apply cubic spline smoothing~\cite{li2010global} to learn each component function. Cubic spline is a piecewise polynomial of degree $ 3 $ with $ C^2 $ continuity interpolating local points in $ \tau $ intervals. Hence, the first and second derivatives of each piecewise term are zero at the endpoints of intervals. We refer to the extended \href{https://arxiv.org/pdf/2206.00667.pdf}{paper} for details of implementation. An ablation study demonstrating the impacts of the hyperparameters $\tau$ and $\lambda$ on the performance of {\framework} is added in the Appendix of the extended \href{https://arxiv.org/pdf/2206.00667.pdf}{paper}. %

\paragraph{Variance and Covariance Computation.} Once each component function $ \function_{\mathbf{a}, \mathbf{S}} $ is learned with $ \textsc{LocalRegression} $ (Line~\ref{algo_line:local_regression_start}--\ref{algo_line:local_regression_end}), we compute variances of the component functions and their covariances using $g_\mathbf{a}(\cdot)$. Since each component function is mean centered (Line~\ref{alg_line:mean_centered}), we compute the variance of $ \function_{\mathbf{a}, \mathbf{S}} $ for the dataset $\mathbf{D}$ as
$  \mathsf{Var}[\function_{\mathbf{a}, \mathbf{S}}] = \sum_{i=1, \mathbf{a}^{(i)} = \mathbf{a}}^n (\function_{\mathbf{a}, \mathbf{S}}(\mathbf{z}_{\mathbf{S}}^{(i)}))^2 $. Hence, variance captures the independent effect of $ \function_{\mathbf{a}, \mathbf{S}} $. Covariance is computed to account for the correlation among features $ \feature $. We compute the covariance of $ \function_{\mathbf{a}, \mathbf{S}} $ with $ \function_{\mathbf{a}} $ on the dataset as
\[
\label{eq:covariance_computation}
	 \mathsf{Cov}[\function_{\mathbf{a}, \mathbf{S}}, \function_{\mathbf{a}}] = \sum_{i=1, \mathbf{a}^{(i)} = \mathbf{a}}^n \function_{\mathbf{a}, \mathbf{S}}(\mathbf{z}_{\mathbf{S}}^{(i)})(g_\mathbf{a}(\mathbf{z}^{(i)}) - \function_{\mathbf{a}, 0}). 
\]
Here, $ g_\mathbf{a}(\cdot) - \function_{\mathbf{a}, 0} $ is the mean centered form of $ \function_{\mathbf{a}} $. Covariance of $ \function_{\mathbf{a}, \mathbf{S}} $ can be both positive and negative depending on whether the features $ \feature_{\mathbf{S}} $ are positively or negatively correlated with $ \function_{\mathbf{a}} $. Specifically, under the set additive model, we obtain $ \mathsf{Cov}[\function_{\mathbf{a}, \mathbf{S}}, \function_{\mathbf{a}}] = \mathsf{Var}[\function_{\mathbf{a}, \mathbf{S}}] + \mathsf{Cov}[\function_{\mathbf{a}, \mathbf{S}}, \sum_{\mathbf{S} \ne \mathbf{S}'} \function_{\mathbf{a}, \mathbf{S}'}] $. Now, we use $ V_{\mathbf{a}, \mathbf{S}} =  \mathsf{Cov}[\function_{\mathbf{a}, \mathbf{S}}, \function_{\mathbf{a}}] $ as the effective variance of $ \feature_{\mathbf{S}} $ for a given sensitive group $ \mathbf{a} $ (Line~\ref{algo_line:covariance_computation_start}--\ref{algo_line:covariance_computation_end}). In Line~\ref{algo_line:fif_computation_start}--\ref{algo_line:fif_computation_end}, we compute $ V_{\mathbf{a}, \mathbf{S}} $ for the most and the least favored groups, and plug them in Eq.~\eqref{eq:fif_decompose} to yield an estimate of FIF of $ \feature_{\mathbf{S}} $.

\begin{proposition}[Time Complexity of \framework] Let $ t $ be the number of iterations of the backfitting algorithm, $ \numfeatures $ be the number of features, $ \lambda $ be the maximum order of intersectional features, and $ s $ be the runtime complexity of the smoothing oracle\footnote{Typically, the runtime complexity of smoothing oracles, particularly of cubic splines, is linear with respect to $ n $, i.e. the number of samples~\cite{toraichi1987computational}.}.	Then, the runtime complexity of Algorithm~\ref{algo:framework} is $ \mathcal{O}\Big(ts\sum_{i=1}^{\lambda} {\numfeatures \choose i}\Big)$.  For example, if we are interested in up to first and second order intersectional features, the runtime complexities are $\mathcal{O}(ts\numfeatures)$ and $\mathcal{O}(ts\numfeatures^2)$, respectively. 
\end{proposition}

\section{Empirical Performance Analysis}\label{sec:experiments}
In this section, we perform an empirical evaluation of {\framework}. Particularly, we discuss the experimental setup, the objectives of experiments, and experimental results. 

\paragraph{Experimental Setup.} We implement a prototype of {\framework} in Python (version $ 3.7.6 $). To estimate FIFs, we leverage and modify the `HDMR' module in SALib library~\cite{Herman2017} based on global sensitivity analysis. In experiments, we consider four widely studied datasets from fairness literature, namely German-credit~\cite{DK2017},
Titanic (\url{https://www.kaggle.com/c/titanic}), COMPAS~\cite{angwin2016machine}, and Adult~\cite{DK2017uci}. We deploy Scikit-learn~\cite{scikit-learn} to learn different classifiers: logistic regression classifier, support vector machine (SVM), neural network, and decision tree with $ 5 $-fold cross-validation. In experiments, we specify {\framework} to compute intersectional influences up to the second order ($ \lambda = 2 $). While applying cubic-spline based local regression in {\framework}, we set $ \tau $, the number of spline intervals to $ 6 $. We compare {\framework} with the existing Shapley-valued based FIF computational framework (\url{https://shorturl.at/iqtuX}), referred as SHAP~\citep{lundberg2020explaining}. For both {\framework} and SHAP, we set a timeout of $ 300 $ seconds for estimating FIFs. In addition, we deploy {\framework} along with a fairness-enhancing algorithm~\citep{kamiran2012data} and a fairness attack~\citep{solans2020poisoning} algorithm, and analyze the effect of these algorithms on the FIFs and the resultant fairness metric. In the following, we discuss the objectives of our empirical study. 

\begin{enumerate}
	\item \textit{Performance:} How \textbf{accurate and computationally efficient} {\framework} and SHAP are in approximating the bias of a classifier based on estimated FIFs?
	\item \textit{Functionality:} How do FIFs estimated by {\framework} and SHAP \textbf{correlate with the impact a fairness intervention} strategy on features?
	\item \textit{Granularity of explanation:} How effective are the \textbf{ intersectional FIFs} in comparison with the \textbf{individual FIFs} while tracing the sources of bias?
	\item \textit{Application:} How do FIFs quantify the impact of \textbf{applying} different fairness enhancing algorithms, i.e. \textbf{affirmative actions}, and fairness attacks, i.e. \textbf{punitive actions}?
\end{enumerate}

In summary, (1) we observe that {\framework} yields \textit{less estimation error} than SHAP while approximating statistical parity using FIFs. {\framework} incurs lower execution time, i.e. \textit{better efficiency in computing individual FIFs} than SHAP while also enabling computation of intersectional FIFs for real-world datasets and classifiers. (2) While considering a fairness intervention, i.e. change in bias due to the omission of a feature, FIFs estimated by {\framework} have \textit{higher correlation with increase/decrease in bias due to  the intervention than SHAP}. %
Thus, {\framework} demonstrates to be \textit{a better choice for identifying features influencing group fairness metrics} than SHAP. 
(3) \textit{By quantifying both individual and intersectional influences of features, \framework{} leads to a more accurate and granular interpretation of the sources of bias}, which is absent in earlier bias explaining methods like SHAP. (4) Finally, as an application of the FIF formulation, {\framework} detects the effects of the affirmative and punitive actions on the bias of a classifier and the corresponding tensions between different subsets of features.  Here, we elaborate on experimental results, and defer additional experiments such as applying {\framework} on other fairness metrics: equalized odds and predictive parity, and an ablation study of hyper-parameters: maximum order of intersectionality  $ \lambda $ and spline intervals $ \tau $ to Appendix of the extended \href{https://arxiv.org/pdf/2206.00667.pdf}{paper}.

\begin{table*}    
	\caption{Median error (over 5-fold cross validation and all combinations of sensitive features) of estimating statistical parity, $ |\mathsf{SP} - \widehat{\mathsf{SP}}| $, using FIFs computed by different methods (columns $ 5 $ to $ 7 $). Best results (lowest error) are in bold color. `{\textemdash}' denotes timeout.}\label{tab:fair_algo_verification}   
	\centering
	\begin{tabular}{lrclrrrr}
		\toprule
		\multirow{2}{*}{Dataset} & \multirow{2}{*}{\shortstack[1]{Dimension\\$ (n, \numfeatures) $}} & \multirow{2}{*}{\shortstack[1]{Max Sensitive\\Features, $ |\sensitive| $}} &\multirow{2}{*}{Classifier} & \multirow{2}{*}{SHAP} & \multicolumn{2}{c}{\framework}\\
		& & & & & $ \lambda = 1 $ & $ \lambda = 2 $\\
		\midrule

\multirow{4}{*}{Titanic} & \multirow{4}{*}{$ (834, 11) $} & \multirow{4}{*}{$ 3$} 
& Logistic Regression &  $ 2.018 $ &  $ 0.218 $ &  $ \mathbf{0.003} $ &  \\ 
& & & SVM &  $ 1.000 $ &  $ 0.137 $ &  $ \mathbf{0.000} $ &  \\ 
& & & Neural Network & \textemdash &  $ 0.215 $ &  $ \mathbf{0.003} $ &  \\ 
& & & Decision Tree &  $ \mathbf{0.018} $ &  $ 0.396 $ &  $ 0.079 $ &  \\ 
\midrule

\multirow{4}{*}{German} & \multirow{4}{*}{$ (417, 23) $} & \multirow{4}{*}{$ 2$} 
& Logistic Regression &  $ 0.361 $ &  $ \mathbf{0.205} $ & \textemdash &  \\ 
& & & SVM &  $ 0.676 $ &  $ \mathbf{0.218} $ & \textemdash &  \\ 
& & & Neural Network & \textemdash &  $ 0.181 $ &  $ \mathbf{0.001} $ &  \\ 
& & & Decision Tree &  $ \mathbf{0.000} $ &  $ 0.262 $ &  $ 0.001 $ &  \\ 
\midrule

\multirow{4}{*}{COMPAS} & \multirow{4}{*}{$ (5771, 8) $} & \multirow{4}{*}{$ 3$} 
& Logistic Regression &  $ 0.468 $ &  $ 0.118 $ &  $ \mathbf{0.056} $ &  \\ 
& & & SVM &  $ 0.360 $ &  $ 0.037 $ &  $ \mathbf{0.020} $ &  \\ 
& & & Neural Network & \textemdash &  $ 0.108 $ &  $ \mathbf{0.053} $ &  \\ 
& & & Decision Tree &  $ \mathbf{0.041} $ &  $ 0.087 $ &  $ 0.055 $ &  \\ 
\midrule

\multirow{4}{*}{Adult} & \multirow{4}{*}{$ (26048, 11) $} & \multirow{4}{*}{$ 3$} 
& Logistic Regression &  $ 2.751 $ &  $ 0.109 $ &  $ \mathbf{0.011} $ &  \\ 
& & & SVM &  $ 0.963 $ &  $ 0.095 $ &  $ \mathbf{0.001} $ &  \\ 
& & & Neural Network & \textemdash &  $ 0.067 $ &  $ \mathbf{0.000} $ &  \\ 
& & & Decision Tree &  $ \mathbf{0.027} $ &  $ 0.146 $ &  $ 0.081 $ &  \\

		\bottomrule
	\end{tabular}
\end{table*}

\subsection{Performance and Functionality of {\framework} in Estimating FIFs}
\paragraph{Accurate Approximation of Bias with FIFs.} We compare {\framework} with SHAP in estimating statistical parity by summing all FIFs, as dictated by the decomposability property (Theorem~\ref{thm:fif_property}). {To our best knowledge, the ground truth of FIF is not known for real-world datasets and classifiers. As such, we cannot compare the accuracy of {\framework} and SHAP directly on the estimated FIFs. Since both methods follow the decomposability property, one way to compare them is to test the accuracy of the sum of estimated FIFs yielding bias/unfairness, as the ground truth of bias of a classifier can be exactly calculated for a given dataset~\cite{aif360-oct-2018}.}
We compute estimation error by taking the absolute difference between the exact and  estimated values of statistical parity, and present median results in Table~\ref{tab:fair_algo_verification}. 

In general, {\framework} achieves less estimation with $ \lambda = 2 $ than with $ \lambda = 1 $ in all datasets and classifiers. This implies that combining intersectional FIFs with individual FIFs captures bias more accurately than the individual FIFs alone. In each dataset, {\framework} ($\lambda = 2$) demonstrates less estimation error than SHAP in all classifiers except in decision trees, denoting that GSA based approach {\framework} is more accurate in approximating group fairness metrics through FIFs than the local explainer SHAP. In decision trees, {\framework}\textemdash which is model-agnostic in methodology\textemdash with $ \lambda = 2 $ often demonstrates a comparable accuracy with SHAP, especially the optimized tree-based explanation counterpart of SHAP~\cite{lundberg2020local2global}. In the context of neural networks, SHAP, particularly Kernel-SHAP, often fails to estimate FIFs within the provided time-limit, while {\framework} with $ \lambda = 2 $ yields highly accurate estimates (median estimation error between $ 0$ to $0.053 $). Therefore, we conclude that \emph{{\framework} is more accurate in estimating statistical parity using FIFs than SHAP.}

\begin{figure}
	\centering
	\subfloat{\includegraphics[scale=0.45]{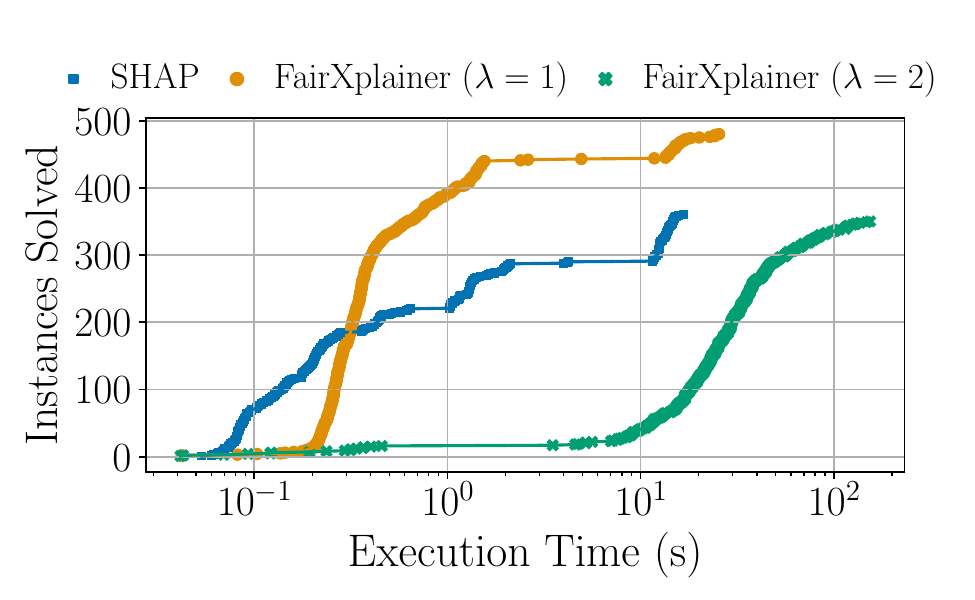}}%
	\caption{Execution time of different methods for estimating FIFs. {\framework} with $ \lambda = 1 $ is more efficient than SHAP, while {\framework} ($ \lambda = 2 $) requires more computational effort.
	}
	\label{fig:execution_time_cactus_plot}
\end{figure}

\paragraph{Execution Time: {\framework} vs.\ SHAP} We compare the execution time of {\framework} vs.\ SHAP in a cactus plot in Figure~\ref{fig:execution_time_cactus_plot}, where a point $ (x, y) $ denotes that a method computes FIFs of $ y $ many fairness instances within $ x $ seconds. We consider $ 480 $ fairness instances from $ 4 $ datasets constituting $ 24 $ distinct combinations of sensitive features, $ 4 $ classifiers, and $ 5 $ cross-validation folds. In Figure~\ref{fig:execution_time_cactus_plot},  {\framework} with $ \lambda = 1 $ is faster than  $ \lambda = 2 $. For example, within $ 10 $ seconds, {\framework} with $ \lambda = 1 $ solves $ 443 $ instances vs.\ $ 41 $ instances with $ \lambda = 2 $.  In addition, {\framework} with $ \lambda = 1 $ solves all $ 480 $ instances compared to $ 360 $ instances solved by SHAP. Thus, {\framework} with $ \lambda = 1 $ demonstrates higher efficiency in estimating individual FIFs compared to SHAP. While estimating intersectional FIFs, {\framework} also demonstrates its practical applicability by solving $ 350 $ instances within  $ 160 $ seconds. \textit{Therefore, {\framework} demonstrates computational efficiency in explaining group fairness metrics of real-world datasets and classifiers.}

\begin{figure}
	\centering%
	\subfloat[COMPAS]{\includegraphics[scale=0.4]{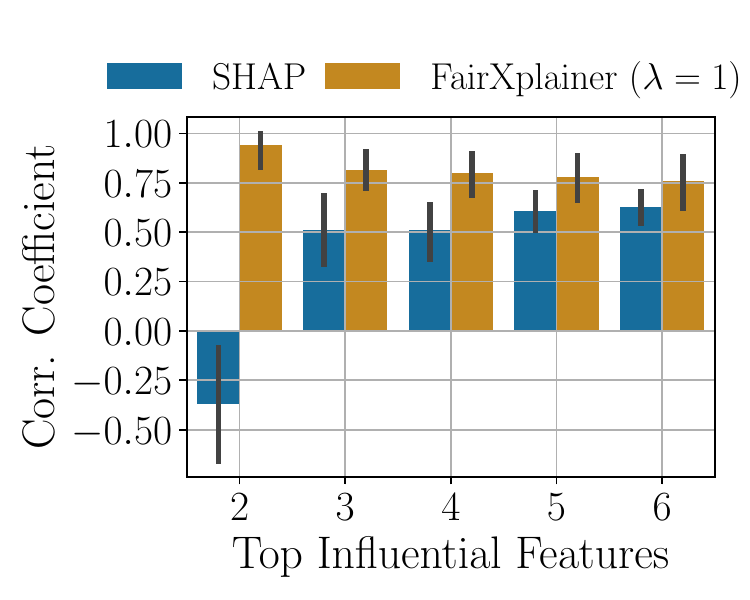}}
	\subfloat[Adult]{\includegraphics[scale=0.4]{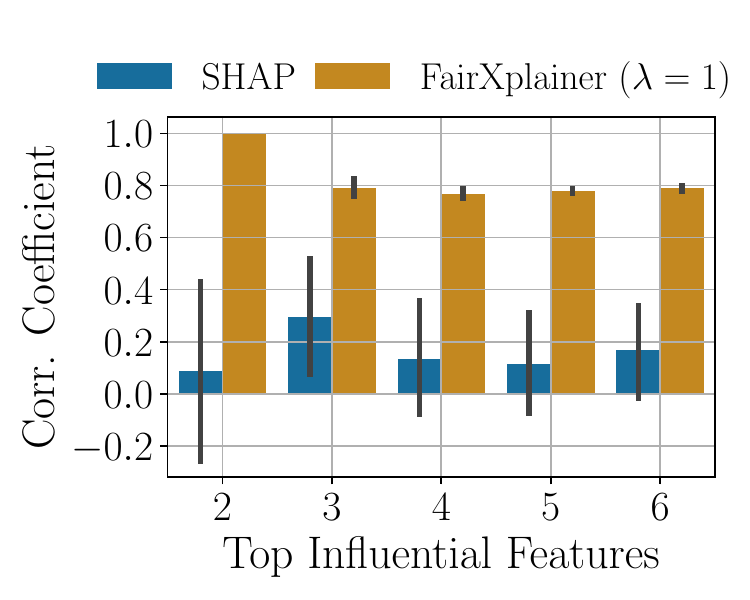}}%
	\caption{Results on fairness intervention on logistic regression classifiers for COMPAS and Adult datasets. Pearson's correlation coefficient between bias-difference due to the intervention and FIFs is higher for top ranked influential features by {\framework} compared to SHAP.}	
	\label{fig:fairness_intervention}
\end{figure}

\paragraph{FIFs under Fairness Intervention} We consider a fairness intervention strategy to hide the impact of an individual feature on a classifier and record the correlation between fairness improvement/reduction of the intervened classifier with the FIF of the feature. Our intervention strategy of modifying the classifier is different than~\cite{datta2016algorithmic}, where the dataset is modified by replacing features with random values. In particular, we intervene a logistic regression classifier by setting the coefficient to zero for the corresponding feature. Intuitively, when the coefficient becomes zero for a feature, the prediction of the classifier may become independent on the feature; thereby, the bias of the classifier may also be independent on the conditional variances of the feature for different sensitive groups.  As a result, a feature with a positive FIF value (i.e. increases bias) is likely to decrease bias under the intervention, and vice versa. In Figure~\ref{fig:fairness_intervention}, we report Pearson's correlation coefficient between the difference in bias (statistical parity) due to intervention and the FIFs of features in COMPAS and Adult datasets, where features are sorted in descending order of their absolute FIFs estimated by {\framework} and SHAP. In {\framework}, the correlation coefficient generally decreases with an increase of top influential features, denoting that features with higher absolute FIFs highly correlate with bias-differences. SHAP, in contrast, demonstrates less correlation, specially for the top most influential features. \emph{Therefore, FIFs estimated by {\framework} shows the potential of being deployed to design improved fairness algorithms in future.}

\begin{figure}
	\centering
	\subfloat[{\framework} ($ \lambda = 1 $)]{\includegraphics[scale=0.38]{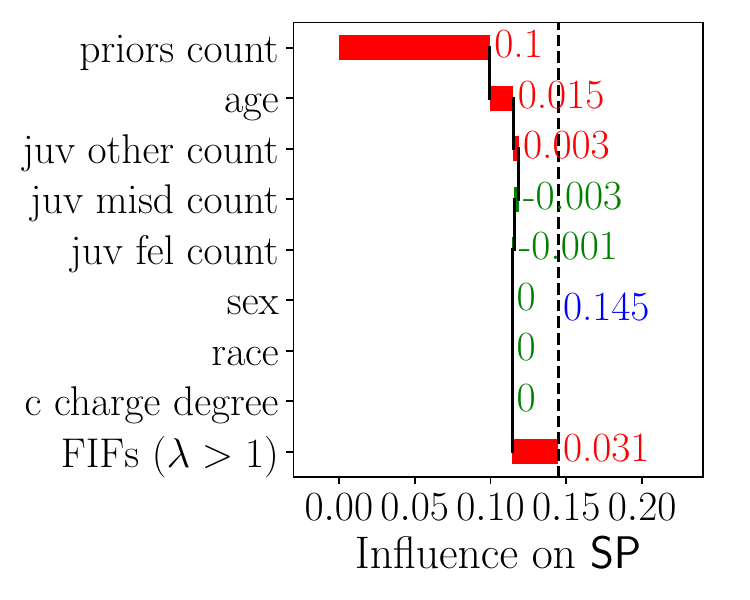}\label{fig:individual_fifs}}
	\subfloat[{\framework} ($ \lambda = 2 $)]{\includegraphics[scale=0.38]{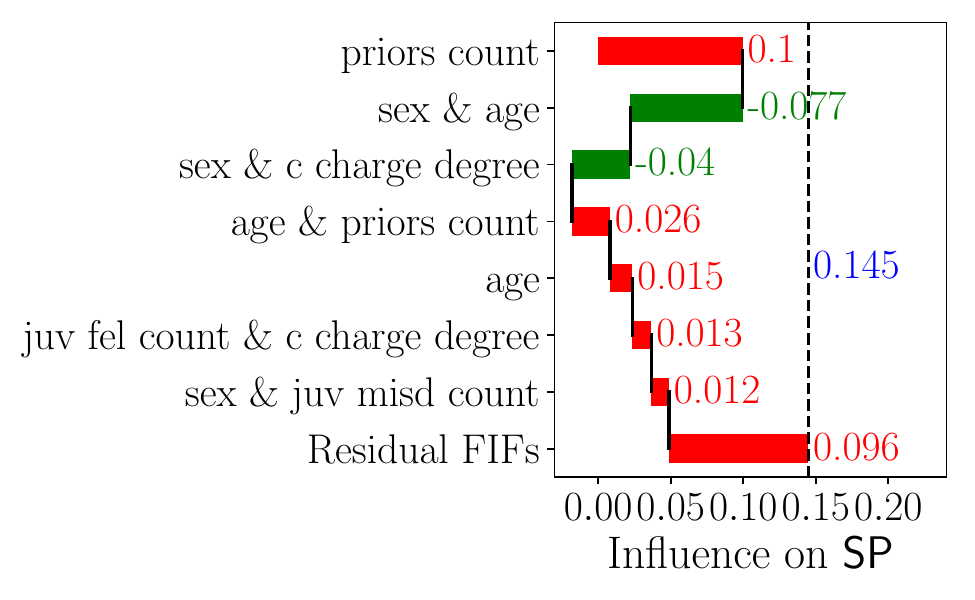}\label{fig:intersectional_fifs}}
	\subfloat[SHAP]{\includegraphics[scale=0.38]{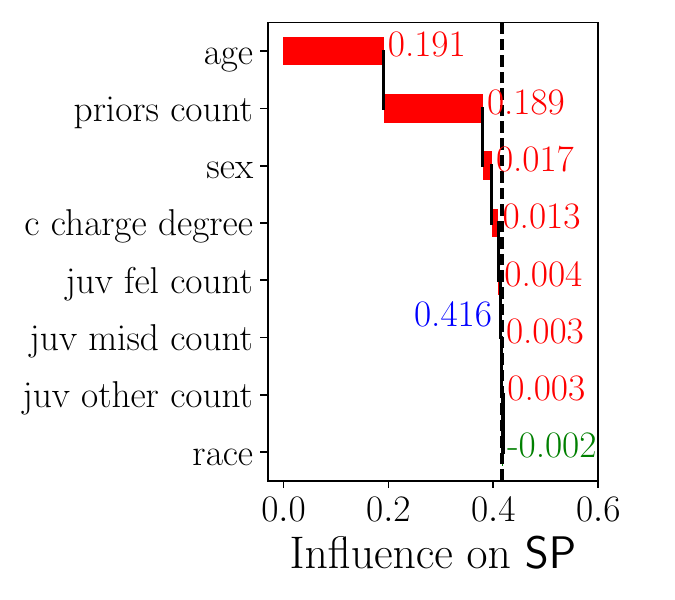}\label{fig:fif_shap}}
	
	\caption{FIFs for COMPAS dataset on explaining statistical parity. Both individual and intersectional FIFs in Figure~\ref{fig:intersectional_fifs} depict sources of bias in detail than individual FIFs alone in Figure~\ref{fig:individual_fifs}. In Figure~\ref{fig:fif_shap}, individual FIFs estimated by SHAP are far from correctly approximating statistical parity (exact value $0.174 $).}
	\label{fig:fif_illustration_compas}
\end{figure}

\subsection{Explainability and Applicability of FIFs}
\paragraph{Individual vs.\ Intersectional FIFs.} Now, we aim to understand the importance of intersectional FIFs over individual FIFs in Figure~\ref{fig:fif_illustration_compas}. We consider COMPAS dataset with race $ \in $ \{Caucasian, non-Caucasian\} as the sensitive feature, and a logistic regression classifier to predict whether a person will re-offend crimes within the next two years. Since the classifier only optimizes training error, it demonstrates statistical parity of $ 0.174 $, i.e. it suggests that a non-Caucasian  has $ 0.174 $ higher probability of re-offending crimes than a Caucasian. Next, we investigate the sources of bias and present individual FIFs in Figure~\ref{fig:individual_fifs}, and both individual and intersectional FIFs in Figure~\ref{fig:intersectional_fifs}. In both figures, we present influential features and their FIFs in the descending order of absolute values. According to {\framework}, the priors count (FIF $ = 0.1 $) dominates in increasing statistical parity\textemdash between Caucasian and non-Caucasian, their prior count demonstrates the maximum difference in the scaled variance of positive prediction. Other non-sensitive features have almost zero FIFs. However, in Figure~\ref{fig:individual_fifs}, higher-order FIFs ($ \lambda > 1 $) increases statistical parity by $ 0.031 $, denoting that the data is correlated and presenting only individual FIFs is not sufficient for understanding the sources of bias. For example, while both sex and age individually demonstrate almost zero influence on bias (Figure~\ref{fig:individual_fifs}), their combined effect and intersectional effects with c-charge degree, priors count, and juvenile miscellaneous count contribute highly on statistical parity. In contrast to {\framework}, SHAP only estimates individual FIFs  (Figure~\ref{fig:fif_shap}) and approximates statistical parity with higher error than {\framework}.  Interestingly, {\framework} yields FIF estimates of the classifier trained on COMPAS dataset that significantly matches with the rule-based classifier extracted from the COMPAS dataset using Certifiably Optimal Rule Lists (CORELS) algorithm~\citep[Figure 3]{rudin19}. From Figure 3 in~\cite{rudin19}, we observe that prior count is the only feature used individually to predict arrests, while age is paired with sex and prior counts respectively. While considering second-order intersectionality, {\framework} ($ \lambda = 2 $) yields priors count, (sex, age), and (age, priors count) as \emph{three of the top four features} that explains the observation of~\cite{rudin19} better than SHAP and {\framework} ($ \lambda = 1 $) considering only individual features. \emph{Therefore, {\framework} demonstrates a clearer understanding on the sources of bias of a classifier by simultaneously quantifying on intersectional influences and individual influences }.  %

\begin{figure}
	\subfloat[Fairness attack (punitive actions)]{\includegraphics[scale=0.38]{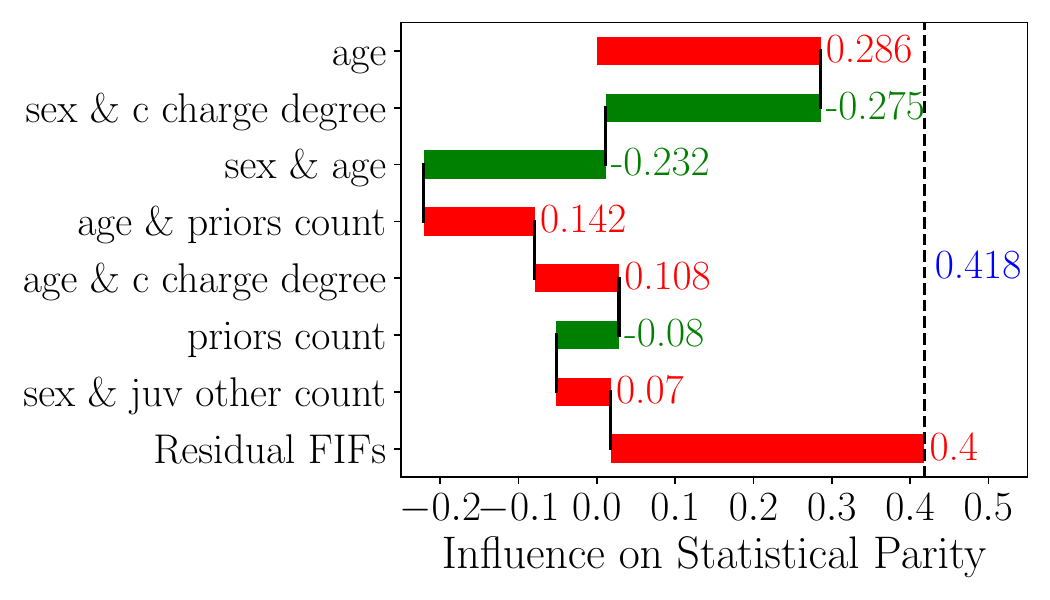}\label{fig:fairness_attack}}\hfil
	\subfloat[Fairness enchancing (affirmative actions)]{\includegraphics[scale=0.38]{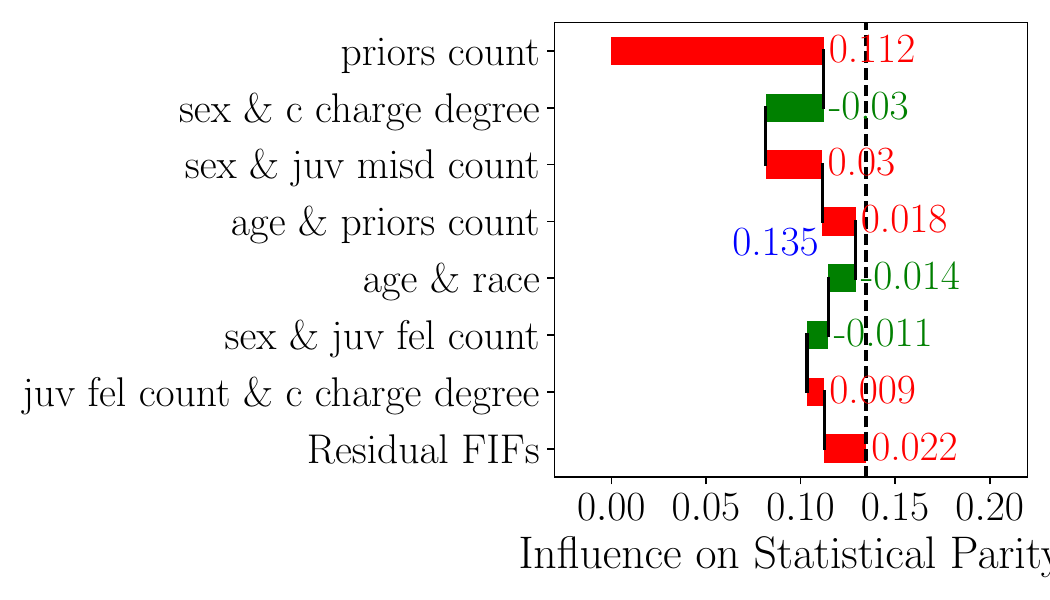}\label{fig:fairness_repair}}
	
	\caption{Effects of a fairness attack~\citep{solans2020poisoning} and a fairness enhancing~\citep{kamiran2012data} algorithms on FIFs.}\label{fig:affirmative_punitive_actions}
\end{figure}

\paragraph{Quantifying Impacts of Affirmative/Punitive Actions.} Continuing on the experiment in Figure~\ref{fig:fif_illustration_compas}, we evaluate the effect of fairness attack and enhancing algorithms on FIFs on COMPAS dataset in Figure~\ref{fig:affirmative_punitive_actions}.  Without applying any fairness algorithm, the statistical parity of the classifier is $ 0.174 $. Applying a data poisoning fairness attack~\citep{solans2020poisoning} increases statistical parity to $ 0.502 $ (approximated as $ 0.418 $ in Figure~\ref{fig:fairness_attack}), whereas  a fairness-enhancing algorithm based on data reweighing~\citep{kamiran2012data} decreases statistical parity to $ 0.151 $ (approximated as $ 0.135 $ in Figure~\ref{fig:fairness_repair}). In Figure~\ref{fig:fairness_attack}, the attack algorithm would be more successful if it could hide the influence of features with positive FIFs, such as priors count and the intersectional effects of age and c-charge degree with sex. In contrast, in Figure~\ref{fig:fairness_repair}, the fairness enhancing algorithm can improve by  further ameliorating the effect features with  negative FIFs, such as priors count. Thus, \textit{{\framework} demonstrates the potential as a dissecting tool to undertake necessary steps to improve or worsen fairness of a classifier.}

\section{Conclusion}

In this paper, we propose the Fairness Influence Function (FIF) to measure the effect of input features on the bias of classifiers on a given dataset. Our approach combines global sensitivity analysis and group-based fairness metrics in machine learning. Thereby, it is natural in our approach to formulate FIF of intersectional features, which together with individual FIFs explains bias with higher granularity. We theoretically analyze the properties of FIFs and provide an algorithm, {\framework}, for estimating FIFs using global variance decomposition and local regression. In experiments, {\framework} estimates individual and intersectional FIFs in real-world datasets and classifiers, approximates bias using FIFs with less estimation error than earlier methods, demonstrates a high correlation between FIFs and fairness interventions, and analyzes the impact of fairness enhancing and attack algorithms on FIFs. The results instantiate {\framework} as a global, granular, and more accurate explanation method to understand the sources of bias. Additionally, the resonance between the rules extracted by CORELS~\cite{rudin19} and the most influential features detected by {\framework} indicates that FIFs can be exploited in future to create an explainable proxy of a biased/unbiased classifier. Also, we aim to extend {\framework} to compute FIFs for complex data, such as image and text, and design algorithms leveraging FIFs to yield unbiased decisions.

\begin{acks}
We acknowledge Reza Shokri and Harold Soh for the elaborate discussion and constructive feedback on the paper. This work is supported in part by National Research Foundation Singapore under its NRF Fellowship Programme [NRF-NRFFAI1-2019-0004], Ministry of Education Singapore Tier 2 grant MOE-T2EP20121-0011, and Ministry of Education Singapore Tier 1 Grant [R-252-000-B59-114].  Bishwamittra Ghosh acknowledges the programme DesCartes and is supported by the National Research Foundation, Prime Minister’s Office, Singapore under its Campus for Research Excellence and Technological Enterprise (CREATE) programme. The early version of this paper is accomplished during the visit of Bishwamittra Ghosh in INRIA, Lille-Nord, France, supported by MOB-LIL-EX (MOBilit\'e- LILle- Excellence) grant. Debabrota Basu is supported by the Inria pilot project "Regalia" and by the French National Research Agency (ANR) through ANR young researcher grant ANR-22-CE23-0003-001 (Project REPUBLIC). The computational work for this paper is performed on resources of the National Supercomputing Centre, Singapore \url{https://www.nscc.sg}.
\end{acks}

\bibliography{main.bib}

\appendix
\appendix
\onecolumn
\section*{Appendix}
\section{Societal Impact}
In this paper, we quantify the influence of input features on the incurred bias/unfairness of the classifier on a dataset. This quantification facilitates our understanding of potential features or subsets of features attributing highly to the bias. This also allows us to understand the effect of the fairness enhancing algorithms in removing bias and their achievement/failure to do so. To the best of our understanding, this paper does not have any negative societal impact.

\section{Proofs of Properties and Implications of FIF}
\begin{reptheorem}{thm:fif_property}
	Let $ f(\alg, \mathbf{D}) $ be the bias/unfairness of the classifier $ \alg $ on dataset $ \mathbf{D} $ according to linear group fairness metrics such as statistical parity. Let $ w_{\mathbf{S}}  $ be the FIF of a subset of  features $ \feature_{\mathbf{S}} $ as defined in Eq.~\eqref{eq:fif_decompose}. 
	\begin{enumerate}
		\item[(a)] \textit{The decomposability property} of FIF states that the sum of FIFs of all subset of  features is equal to the bias of the classifier. 
		\begin{align}
		\sum_{\mathbf{S} \subseteq [\numfeatures] } w_{\mathbf{S}} = f(\alg, \mathbf{D})
		\end{align}
		\item[(b)] \textit{The symmetry property} states that  two features $ Z_i $ and $ Z_j $ are equivalent based on FIF if the sum of corresponding individual influences and the intersectional influences with all other features are the same. Mathematically,
		\begin{align}
		\sum_{\mathbf{S}'' \subseteq [m]\setminus \{i,j\}}w_{\mathbf{S}''\cup \{i\}} = \sum_{\mathbf{S}'' \subseteq [m]\setminus\{i,j\}}w_{\mathbf{S}''\cup \{j\}}  
		\end{align}
		if $\sum_{\mathbf{S}' \subseteq \mathbf{S}\cup \{i\}} w_{\mathbf{S}'} = 		\sum_{\mathbf{S}' \subseteq \mathbf{S} \cup \{j\}} w_{\mathbf{S}'}$ for every non-empty subset $ \mathbf{S} $ of $ [\numfeatures] $ containing neither $ i $ nor $ j $. 
		\item[(c)] \textit{The null property} of FIF states that feature $ X_i $ is a dummy or neutral feature if sum of its individual influence and the intersectional influences with all other features is zero. Mathematically,
		\begin{align}
		\sum_{\mathbf{S}'' \subseteq [m]\setminus\{i\}}w_{\mathbf{S}''\cup \{i\}} = 0 	 
		\end{align}
		if	$\sum_{\mathbf{S}' \subseteq \mathbf{S} \cup \{i\}} w_{\mathbf{S}'} = 		\sum_{\mathbf{S}' \subseteq \mathbf{S}} w_{\mathbf{S}'}$ for every non-empty subset $ \mathbf{S} $ of $ [\numfeatures] $ that does not contain $ i $.
	\end{enumerate}
\end{reptheorem}

\begin{proof}
	(a) The decomposability property of FIF is based on GSA, where the total variance is decomposed to the variances of individual and intersectional inputs. 
	\begin{align*}
	\sum_{\mathbf{S} \subseteq [\numfeatures]} w_{\mathbf{S}} &= \sum_{\mathbf{S} \subseteq [\numfeatures]} \frac{V_{\mathbf{a}_{\max}, \mathbf{S}}}{ \Pr[\widehat{Y} = 0 \mid  \sensitive = \mathbf{a}_{\max}]} - \frac{V_{\mathbf{a}_{\min}, \mathbf{S}} }{\Pr[\widehat{Y} = 0 \mid  \sensitive = \mathbf{a}_{\min}]}\\
	&= \frac{\sum_{\mathbf{S} \subseteq [\numfeatures]} V_{\mathbf{a}_{\max}, \mathbf{S}}}{ \Pr[\widehat{Y} = 0 \mid  \sensitive = \mathbf{a}_{\max}]} - \frac{\sum_{\mathbf{S} \subseteq [\numfeatures]} V_{\mathbf{a}_{\min}, \mathbf{S}} }{\Pr[\widehat{Y} = 0 \mid  \sensitive = \mathbf{a}_{\min}]}\\
	&= \frac{V_{\mathbf{a}_{\max}}}{ \Pr[\widehat{Y} = 0 \mid  \sensitive = \mathbf{a}_{\max}]} - \frac{V_{\mathbf{a}_{\min}} }{\Pr[\widehat{Y} = 0 \mid  \sensitive = \mathbf{a}_{\min}]}\\
	&=\frac{\mathsf{Var}[\widehat{Y} = 1\mid\sensitive = \mathbf{a}_{\max}]}{\Pr[\widehat{Y} = 0 \mid  \sensitive = \mathbf{a}_{\max}]} - \frac{\mathsf{Var}[\widehat{Y} = 1\mid\sensitive = \mathbf{a}_{\min}] }{\Pr[\widehat{Y} = 0 \mid  \sensitive = \mathbf{a}_{\min}]}\\
	&= f_{\mathsf{SP}}(\alg, \mathbf{D})
	\end{align*}
	Thus, applying Lemma~\ref{lm:sp_var_relation}, we prove the decomposability property of FIF for statistical parity.
	
	(c) We observe that 
	\begin{align}\label{eq:sum_decomp}
		\sum_{\mathbf{S}' \subseteq \mathbf{S}\cup \{i\}} w_{\mathbf{S}'} = \sum_{\mathbf{S}' \subseteq \mathbf{S}} w_{\mathbf{S}'} + \sum_{\mathbf{S}'' \subseteq \mathbf{S}} w_{\mathbf{S}''\cup \{i\}}.
	\end{align}
	This means that we can decompose the sum of FIFs of all the subsets of $\mathbf{S}\cup \{i\}$ into two non-overlapping sums: the subsets that include $i$ and the subsets that do not include.
	
	Since we assume that for $i$,  $\sum_{\mathbf{S}' \subseteq \mathbf{S}\cup \{i\}} w_{\mathbf{S}'} = \sum_{\mathbf{S}' \subseteq \mathbf{S}} w_{\mathbf{S}'}$ holds true, it implies
		\begin{align*}
	\sum_{\mathbf{S}'' \subseteq \mathbf{S}} w_{\mathbf{S}''\cup \{i\}} = 0.
	\end{align*}
	Now, considering $\mathbf{S} = [m]\setminus \{i\}$, i.e. the set of all features  except $i$, concludes the proof.
	
	(b) Proof of the symmetry property follows similar decomposition of the sum of FIFs as Equation~\eqref{eq:sum_decomp}.
\end{proof}

\begin{repproposition}{prop:neg_fif}
	When $ w_{\mathbf{S}} < 0 $, i.e. features $ \feature_{\mathbf{S}} $ decrease bias, the decomposed variance of CPPs w.r.t.\ $ \feature_{\mathbf{S}} $ follows $ V_{\mathbf{a}_{\max}, \mathbf{S}} < V_{\mathbf{a}_{\min}, \mathbf{S}}  $.
\end{repproposition}

\begin{proof}
	When $ w_{\mathbf{S}} < 0 $,
	\begin{align*}
	&\frac{V_{\mathbf{a}_{\max}, \mathbf{S}}}{ \Pr[\widehat{Y} = 0 \mid  \sensitive = \mathbf{a}_{\max}]} - \frac{V_{\mathbf{a}_{\min}, \mathbf{S}} }{\Pr[\widehat{Y} = 0 \mid  \sensitive = \mathbf{a}_{\min}]} < 0 \\
	\implies&\frac{V_{\mathbf{a}_{\max}, \mathbf{S}}}{ \Pr[\widehat{Y} = 0 \mid  \sensitive = \mathbf{a}_{\max}]} < \frac{V_{\mathbf{a}_{\min}, \mathbf{S}} }{\Pr[\widehat{Y} = 0 \mid  \sensitive = \mathbf{a}_{\min}]} \\
	\implies&\frac{V_{\mathbf{a}_{\max}, \mathbf{S}}}{V_{\mathbf{a}_{\min}, \mathbf{S}}} < \frac{\Pr[\widehat{Y} = 0 \mid  \sensitive = \mathbf{a}_{\max}]}{\Pr[\widehat{Y} = 0 \mid  \sensitive = \mathbf{a}_{\min}]} \le 1 \\
	\implies&\frac{V_{\mathbf{a}_{\max}, \mathbf{S}}}{V_{\mathbf{a}_{\min}, \mathbf{S}}} < 1 \\ 
	\implies&V_{\mathbf{a}_{\max}, \mathbf{S}} < V_{\mathbf{a}_{\min}, \mathbf{S}}
	\end{align*}
\end{proof}

\begin{repproposition}{prop:pos_fif}
	{If the decomposed variance of CPPs w.r.t.\ $ \feature_{\mathbf{S}} $ satisfies $ V_{\mathbf{a}_{\max}, \mathbf{S}} > {V_{\mathbf{a}_{\min}, \mathbf{S}}} $, the corresponding FIF $ w_{\mathbf{S}} > 0  $, i.e. features $ \feature_{\mathbf{S}} $ increase bias.}
\end{repproposition}
\begin{proof}\
	Since $	V_{\mathbf{a}_{\max}, \mathbf{S}} > V_{\mathbf{a}_{\min}, \mathbf{S}}$, we obtain $
	\frac{V_{\mathbf{a}_{\max}, \mathbf{S}}}{V_{\mathbf{a}_{\min}, \mathbf{S}}} > 1.$
Since $\mathbf{a}_{\max}$ and $\mathbf{a}_{\min}$ are the most and least favored groups respectively, the probability of yielding a positive prediction is greater or equal for $\mathbf{a}_{\max}$ than $\mathbf{a}_{\min}$. Thus, ${\Pr[\widehat{Y} = 0 \mid  \sensitive = \mathbf{a}_{\max}]} \leq {\Pr[\widehat{Y} = 0 \mid  \sensitive = \mathbf{a}_{\min}]}$, which implies that $\frac{\Pr[\widehat{Y} = 0 \mid  \sensitive = \mathbf{a}_{\max}]}{\Pr[\widehat{Y} = 0 \mid  \sensitive = \mathbf{a}_{\min}]} \leq 1$.

Combining both the observations, we obtain
\begin{align*}
&\frac{V_{\mathbf{a}_{\max}, \mathbf{S}}}{V_{\mathbf{a}_{\min}, \mathbf{S}}} > \frac{\Pr[\widehat{Y} = 0 \mid  \sensitive = \mathbf{a}_{\max}]}{\Pr[\widehat{Y} = 0 \mid  \sensitive = \mathbf{a}_{\min}]}\\
\implies &\frac{V_{\mathbf{a}_{\max}, \mathbf{S}}}{\Pr[\widehat{Y} = 0 \mid  \sensitive = \mathbf{a}_{\max}]} > \frac{V_{\mathbf{a}_{\min}, \mathbf{S}}}{\Pr[\widehat{Y} = 0 \mid  \sensitive = \mathbf{a}_{\min}]}\\
\implies &\frac{V_{\mathbf{a}_{\max}, \mathbf{S}}}{\Pr[\widehat{Y} = 0 \mid  \sensitive = \mathbf{a}_{\max}]} - \frac{V_{\mathbf{a}_{\min}, \mathbf{S}}}{\Pr[\widehat{Y} = 0 \mid  \sensitive = \mathbf{a}_{\min}]} > 0\\
\implies &w_{\mathbf{S}} > 0.
\end{align*}
\end{proof}

\section{A Smoothing Operator $\textsc{Smooth}$: Cubic Splines}
\label{sec:smoothing} 
In the $\textsc{LocalRegression}$ module of \framework{} (Line~\ref{algo_line:local_regression_start}--\ref{algo_line:local_regression_end}, Algorithm~\ref{algo:framework}), we use a smoothing operator $\textsc{Smooth}$ (Line~\ref{alg_line:backfitting_step}). In our experiments, \textit{we use cubic splines as the smoothing operator}. Here, we elucidate the technical details of cubic splines.

In interpolation problems, a B-spline of order $ n $ is traditionally used to smoothen the intersection of piecewise interpolators~\cite{schumaker2007spline}. A B-spline of degree $n$ is a piecewise polynomial of degree $ n - 1 $ defined over a variable $ Z $. Each piece-wise term is computed on local points and is aggregated as a global curve smoothly fitting the data. The values of $ Z $ where the polynomial pieces meet together are called knots, and are denoted by $\{ \dots, t_0, t_1, t_2, \dots\} $. 

Let $ B_{r, n}(Z) $ denote the basis function for a B-spline of order $ n $, and $ r $ is the index of the knot vector. According to Carl de Boor~\cite{de1971subroutine}, $ B_{r,1}(Z) $, for $ n = 1 $, is defined as
\begin{align*}
B_{r,1}(Z) = 
\begin{cases}
0 &\text{ if } Z < t_r \text{ or } Z \ge t_{r+1},\\
1 &\text{ otherwise}
\end{cases}
\end{align*}

This definition satisfies $ \sum_i B_{r, 1}(Z) = 1 $. The higher order basis functions are defined recursively as
\begin{align*}
	B_{r, n + 1}(Z) = p_{r, n}(Z)B_{r, n}(Z) + (1 - p_{r + 1, n}(Z))B_{r + 1, n}(Z),
\end{align*}
where 
\begin{align*}
p_{r,n}(Z) = 
\begin{cases}
\frac{Z - t_r}{t_{r + n} - t_r} &\text{ if } t_{r + n} \ne t_r,\\
0 &\text{ otherwise.}
\end{cases}
\end{align*}

In this paper, we consider cubic splines with the basis function $ B_{r,4}(Z) $ that constitutes a B-spline of degree $ 3 $. This polynomial has $ C^2 $ continuity, i.e. for each piecewise term, derivatives up to the second order are zero at the endpoints of each interval in the knot vector. We estimate component functions $ \function_{\mathbf{a}, \mathbf{S}} $'s with the basis function $ B_{r,4}(\mathbf{Z})  $ of cubic splines~\cite{li2010global}, as shown in Equation~\eqref{eq:cubic_splines}.  
\begin{align}\label{eq:cubic_splines}
\begin{split}
&\function_{\mathbf{a}, \{i\}} (\mathbf{Z}_{\{i\}}) \approx \sum_{r=-1}^{\tau+1}\alpha_r^iB_{r,n}(\mathbf{Z}_{\{i\}})\\
&\function_{\mathbf{a}, \{i, j\}} (\mathbf{Z}_{\{i, j\}}) \approx \sum_{p=-1}^{\tau+1}\sum_{q=-1}^{\tau+1}\beta_{pq}^{ij}B_p(\mathbf{Z}_{\{i\}})B_q(\mathbf{Z}_{\{j\}})\\
&\function_{\mathbf{a}, \{i, j, k\}} (\mathbf{Z}_{\{i, j, k\}}) \approx \sum_{p=-1}^{\tau+1}\sum_{q=-1}^{\tau+1}\sum_{r=-1}^{\tau+1}\gamma_{pqr}^{ijk}B_p(\mathbf{Z}_{\{i\}})B_q(\mathbf{Z}_{\{j\}})B_{r,n}(\mathbf{Z}_{\{j\}})
\end{split}
\end{align}

Here, $ \tau $ is the number of knots, also called spline intervals. We learn the coefficients $ \alpha, \beta, \gamma $ using the backfitting algorithm (Line~\ref{algo_line:local_regression_start}--\ref{algo_line:local_regression_end}, Algorithm~\ref{algo:framework}). 

$\tau$ is a hyper-parameter that influences the accuracy of the local regression and thus, the FIFs. We perform an ablation study to explicate the impact of $\tau$ on the performance of {\framework} in Appendix~\ref{sec:additional_experiments}.

\section{Computing FIFs for Equalized Odds and Predictive Parity}\label{app:pp}
Due to brevity of space, we elaborate definition of statistical parity and corresponding methodology to compute FIFs in the main text. Here, we provide definition of other group-based fairness metrics~\citep{verma2018fairness}: equalized odds and predictive parity. We also explain the methodology to use {\framework} in order to compute FIFs corresponding to these metrics.

Following the classification setting and notations described in Section~\ref{sec:preliminaries}, here we consider a binary classifier trained on dataset $\mathbf{D}\triangleq \{(\mathbf{x}^{(i)}, \mathbf{a}^{(i)}, y^{(i)})\}_{i=1}^n$ as $\alg: (\nonsensitive, \sensitive) \rightarrow \widehat{Y} $. $\widehat{Y} \in \{0,1\}$ and ${Y} \in \{0,1\}$ represents the predicted class and the true class for a data point $ (\nonsensitive, \sensitive) $ with sensitive features $\sensitive$ and non-sensitive features $\nonsensitive$.

\subsection{Equalized Odds} 

\textit{Equalized Odds} ($ \mathsf{EO} $)~\citep{hardt2016equality}: \textit{Separation} measuring group fairness metrics such as equalized odds constrain that the predicted class $ \widehat{Y} $ is independent of $ \sensitive $ given the true class $ Y $.  Formally, for $ Y \in \{0,1\} $, equalized odds is defined as
\[ f_{\mathsf{EO}}(\alg, \mathbf{D}) \triangleq  \max_{y} \Big(\max_{\mathbf{a}}\Pr[\widehat{Y} =1 \mid \mathbf{A} = \mathbf{a}, Y = y] - \min_{\mathbf{a}}\Pr[\widehat{Y} =1 \mid \mathbf{A} = \mathbf{a}, Y = y]\Big).
\] 

Equalized odds intuitively implies the maximum statistical parity conditioned on $ Y $.

\paragraph{Extension of {\framework} to Equalized Odds.}  For equalized odds, we deploy {\framework} twice, one for computing FIFs on a subset of data points in the dataset where $ Y = 1 $ and another on data points with $ Y = 0 $. Since, the maximum of the sum of FIFs between $ Y = 1 $ and $ Y = 0 $ is the equalized odds of the classifier, we finally report FIFs of features corresponding to the maximum sum of FIFs between $ Y \in \{0, 1\} $.

\subsection{Predictive Parity} 

\textit{Sufficiency} measuring group fairness metrics such as Predictive Parity ($ \mathsf{PP} $) constrain that the ground class $ Y $ is independent of $ \sensitive $ given the prediction $ \widehat{Y} $. Formally, 
\begin{align*}
	f_{\mathsf{PP}}(\alg, \mathbf{D})  
	\triangleq \max_{y} \Big(\max_{\mathbf{a}}\Pr[Y =1 \mid \mathbf{A} = \mathbf{a}, \widehat{Y} = y] - \min_{\mathbf{a}}\Pr[Y =1 \mid \mathbf{A} = \mathbf{a}, \widehat{Y} = y]\Big).
\end{align*} 

\paragraph{Extension of {\framework} to Predictive Parity.}  To compute FIFs for predictive parity, we condition the dataset by the predicted class $ \widehat{Y} $ and separate into two sub-datasets: $ \widehat{Y} = 1 $ and  $ \widehat{Y} = 0 $. For each sub-dataset, we deploy $ \framework $ by setting the ground-truth class $ Y $ as label. This contrasts the computation for  statistical parity and equalized odds, where the predicted class $ \widehat{Y} $ is considered as label. Finally, the maximum of the sum of FIFs between two sub-datasets for $ \widehat{Y} = 1 $ and  $ \widehat{Y} = 0 $ measures the predictive parity. Similar to equalized odds, FIFs achieving the greatest sum of FIFs for $ \widehat{Y} \in \{0, 1\} $ are the reported FIFs for the predictive parity of the classifier.

\begin{table}
	
	\caption{Median estimation error (over 5-fold cross validation and all combinations of sensitive features) of equalized odds (columns $ 5 $ to $ 7 $) and predictive parity (columns $ 8 $ and $ 9 $) in terms of estimated FIFs by different methods. Best results (lowest error) are in bold color. `{\textemdash}' denotes timeout. SHAP cannot estimate FIFs for predictive parity due to limited methodology.}
	\label{tab:accuracy_eo_pp}
	\footnotesize
	\centering
	\begin{tabular}{lrclrrr|rrr}
		\toprule
		\multirow{3}{*}{Dataset} & \multirow{3}{*}{\shortstack[1]{Dimension\\$ (n, \numfeatures) $}} & \multirow{3}{*}{\shortstack[1]{Max $ |\sensitive| $}} &\multirow{3}{*}{Classifier} & \multicolumn{3}{c|}{Equalized Odds} & \multicolumn{3}{c}{Predictive Parity} \\
		& & & & \multirow{2}{*}{SHAP} & \multicolumn{2}{c|}{\framework} &  \multicolumn{2}{c}{\framework}\\
		& & & & & $ \lambda = 1 $ & $ \lambda = 2 $ & $ \lambda = 1 $ & $ \lambda = 2 $ \\
		\midrule

		\multirow{4}{*}{Titanic} & \multirow{4}{*}{$ (834, 11) $} & \multirow{4}{*}{$ 3$} 
		& Logistic Regression & $ 1.697 $  & $ \mathbf{0.000} $  & $ \mathbf{0.000} $  & $ 0.251 $  & $ \mathbf{0.148} $  \\ 
		& & & SVM & $ 1.000 $  & $ \mathbf{0.000} $  & $ \mathbf{0.000} $  & $ 0.092 $  & $ \mathbf{0.045} $  \\ 
		& & & Neural Network & \textemdash  & $ \mathbf{0.000} $  & $ \mathbf{0.000} $  & $ 0.349 $  & $ \mathbf{0.171} $  \\ 
		& & & Decision Tree & $ 0.074 $  & $ 0.185 $  & $ \mathbf{0.059} $  & $ \mathbf{0.097} $  & $ \mathbf{0.097} $  \\ 
		\midrule
		
		\multirow{4}{*}{German} & \multirow{4}{*}{$ (417, 23) $} & \multirow{4}{*}{$ 2$} 
		& Logistic Regression & $ 0.382 $  & $ 0.109 $  & $ \mathbf{0.001} $  & $ 0.075 $  & $ \mathbf{0.001} $  \\ 
		& & & SVM & $ 0.435 $  & $ \mathbf{0.082} $  & \textemdash  & $ 0.060 $  & $ \mathbf{0.001} $  \\ 
		& & & Neural Network & \textemdash  & $ 0.149 $  & $ \mathbf{0.000} $  & $ 0.184 $  & $ \mathbf{0.000} $  \\ 
		& & & Decision Tree & $ \mathbf{0.000} $  & $ \mathbf{0.000} $  & $ \mathbf{0.000} $  & $ \mathbf{0.000} $  & $ \mathbf{0.000} $  \\ 
		\midrule
		
		\multirow{4}{*}{COMPAS} & \multirow{4}{*}{$ (5771, 8) $} & \multirow{4}{*}{$ 3$} 
		& Logistic Regression & $ 0.380 $  & $ 0.167 $  & $ \mathbf{0.071} $  & $ \mathbf{0.201} $  & $ 0.214 $  \\ 
		& & & SVM & $ 0.481 $  & $ 0.043 $  & $ \mathbf{0.024} $  & $ 0.124 $  & $ \mathbf{0.117} $  \\ 
		& & & Neural Network & \textemdash  & $ 0.143 $  & $ \mathbf{0.055} $  & $ \mathbf{0.078} $  & $ 0.129 $  \\ 
		& & & Decision Tree & $ 0.071 $  & $ 0.069 $  & $ \mathbf{0.031} $  & $ 0.348 $  & $ \mathbf{0.340} $  \\ 
		\midrule
		
		\multirow{4}{*}{Adult} & \multirow{4}{*}{$ (26048, 11) $} & \multirow{4}{*}{$ 3$} 
		& Logistic Regression & $ 1.647 $  & $ 0.186 $  & $ \mathbf{0.013} $  & $ 0.090 $  & $ \mathbf{0.002} $  \\ 
		& & & SVM & $ 0.703 $  & $ 0.081 $  & $ \mathbf{0.001} $  & $ 0.109 $  & $ \mathbf{0.002} $  \\ 
		& & & Neural Network & \textemdash  & $ 0.077 $  & $ \mathbf{0.000} $  & $ 0.091 $  & $ \mathbf{0.002} $  \\ 
		& & & Decision Tree & $ \mathbf{0.062} $  & $ 0.263 $  & $ 0.190 $  & $ 0.216 $  & $ \mathbf{0.203} $  \\

		\bottomrule
	\end{tabular}
\end{table}

\section{Experimental Evaluations}\label{sec:additional_experiments}
\subsection{Experimental Setup}
We perform experiments on a Red Hat Enterprise Linux Server release 6.10 (Santiago) that has an $\text{E}5-2690 \text{ v}3$ CPU and 16GB of RAM. With the aim of computing FIFs for any classifier, we do not adjust the classifier's hyper-parameters during training. Instead, we utilize the default hyper-parameters provided by Scikit-learn~\cite{scikit-learn}. For equalized odds and predictive parity, {\framework}  (similarly SHAP) is deployed twice. Hence, we double the time limit, $ 2*300 = 600 $ seconds.

\subsection{Accurate Approximation of Equalized Odds and Predictive Parity using FIFs.}
The accuracy of approximating equalized odds and predictive parity for {\framework} and SHAP is compared in Table~\ref{tab:accuracy_eo_pp}. {\framework} shows lower estimation error compared to SHAP, especially when $\lambda=2$. SHAP is unable to explain predictive parity as predictive parity relies on the ground label $Y$, which is not available for randomly generated data points by SHAP for estimating local explanations.

\begin{figure}
	\centering
	\subfloat[Equalized Odds]{\includegraphics[scale=0.4]{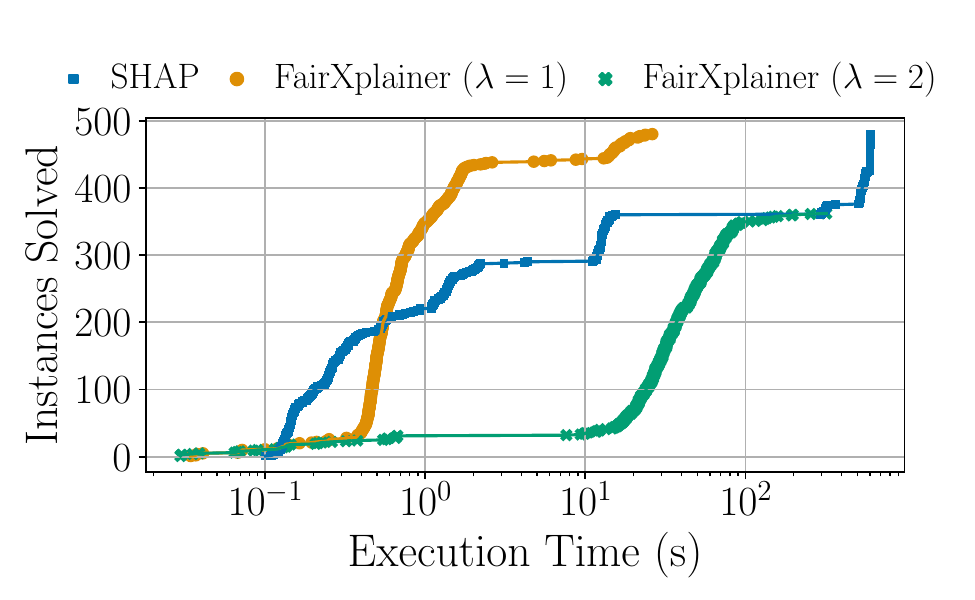}} 
	\subfloat[Predictive Parity]{\includegraphics[scale=0.4]{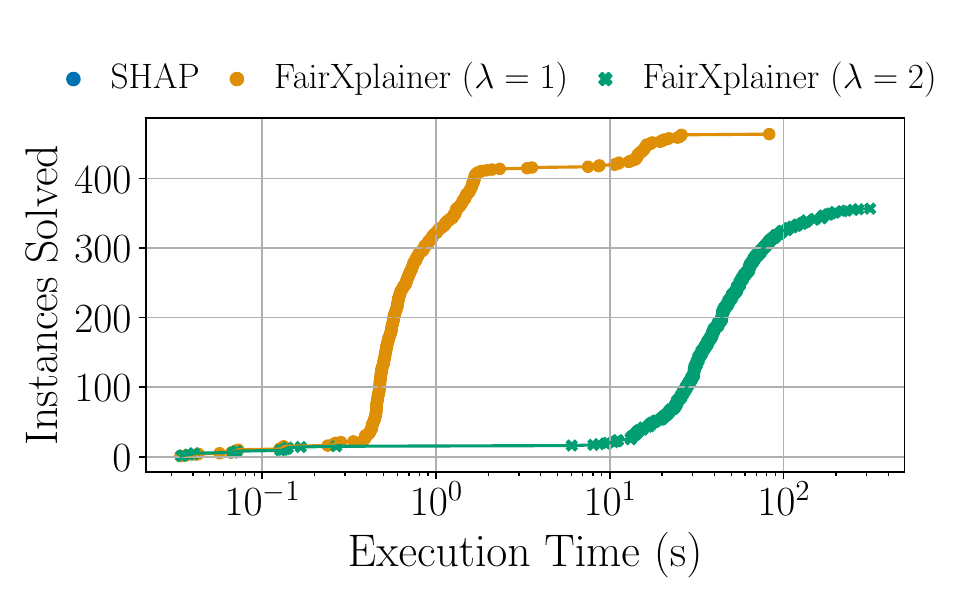}}
	\caption{Execution time of different methods for estimating FIFs for equalized odds and predictive parity. {\framework} with $ \lambda = 1 $ is more efficient than SHAP, while {\framework} ($ \lambda = 2 $) requires more computational effort. SHAP cannot explain predictive parity. 
	}
	\label{fig:execution_time_cactus_plot_eo_pp}
\end{figure}

\subsection{Execution Time for FIF Estimation of Equalized Odds and Predictive Parity}
In Figure~\ref{fig:execution_time_cactus_plot_eo_pp}, we demonstrate the execution time of different methods in estimating FIFs of equalized odds and predictive parity in cactus plots. {\framework} with $ \lambda = 1 $ is more efficient than $ \lambda = 2 $ and solves all $ 480 $ fairness instances with at least one order of magnitude less execution time. Compared to {\framework} ($ \lambda = 1 $), SHAP demonstrates less computational efficiency.

\begin{figure}
	\subfloat[Adult]{\includegraphics[scale=0.4]{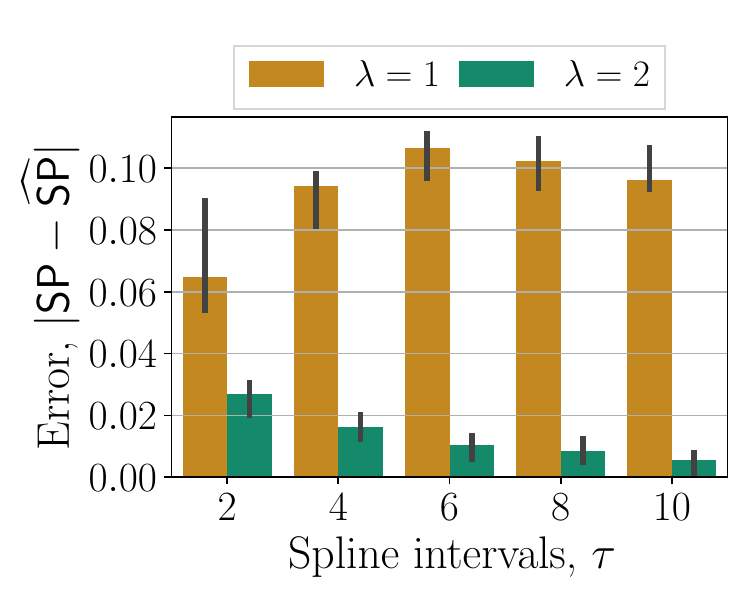}}
	\subfloat[Adult]{\includegraphics[scale=0.4]{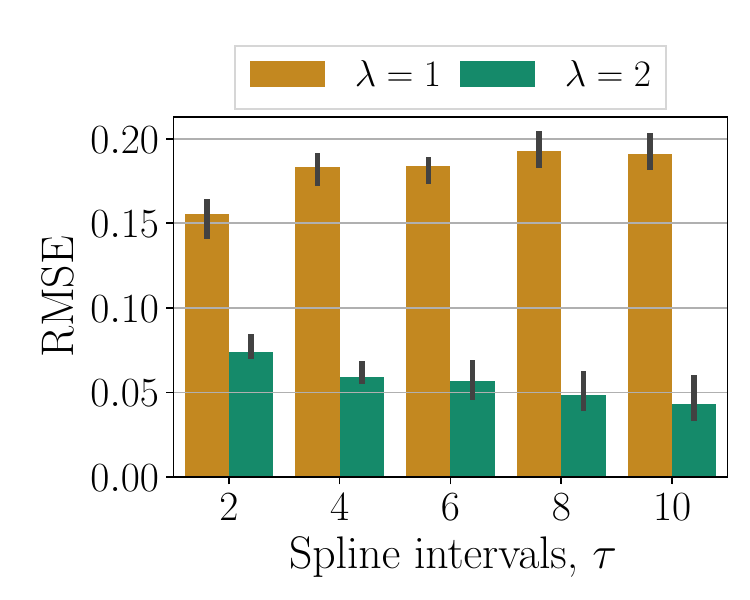}}
	\subfloat[Adult]{\includegraphics[scale=0.4]{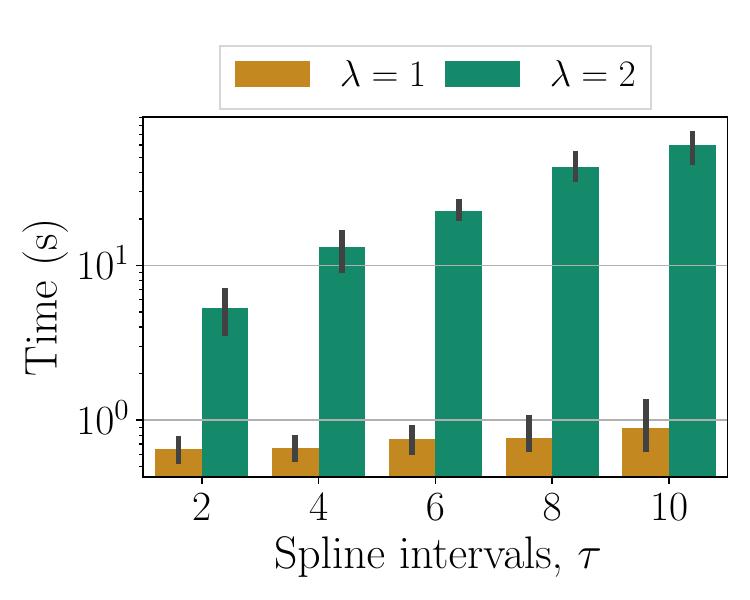}}
	
	\subfloat[COMPAS]{\includegraphics[scale=0.4]{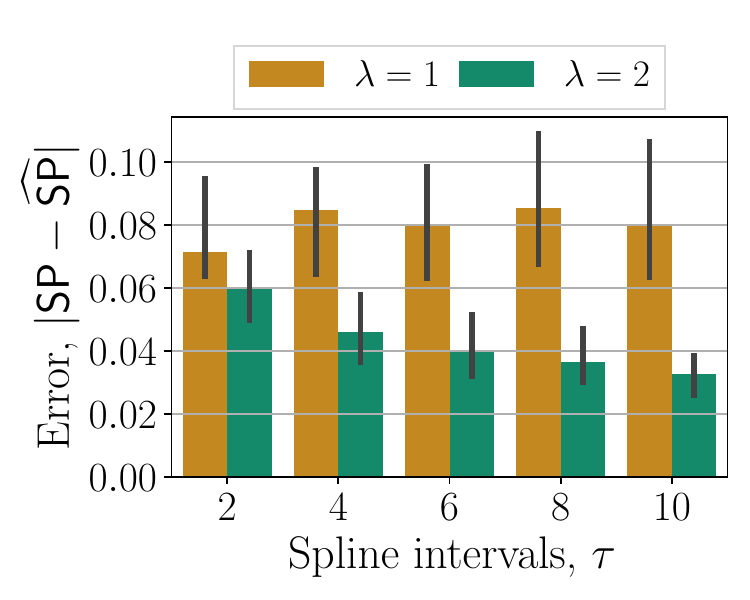}}
	\subfloat[COMPAS]{\includegraphics[scale=0.4]{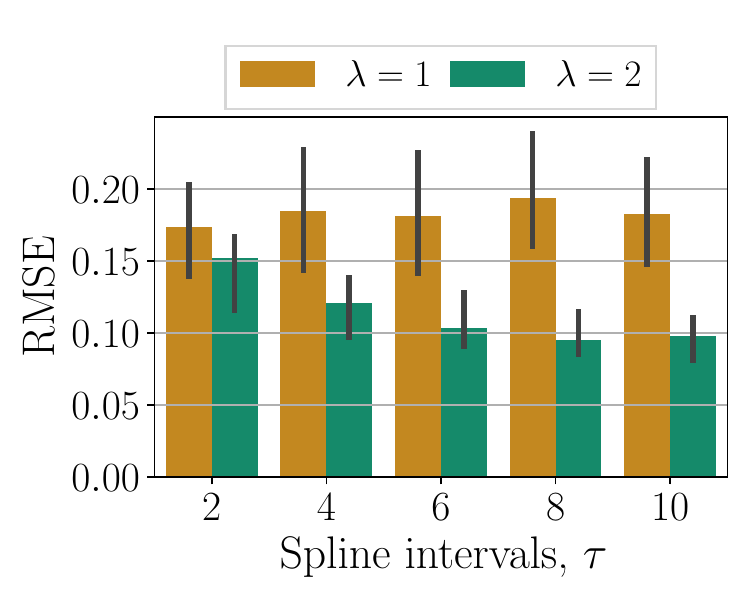}}
	\subfloat[COMPAS]{\includegraphics[scale=0.4]{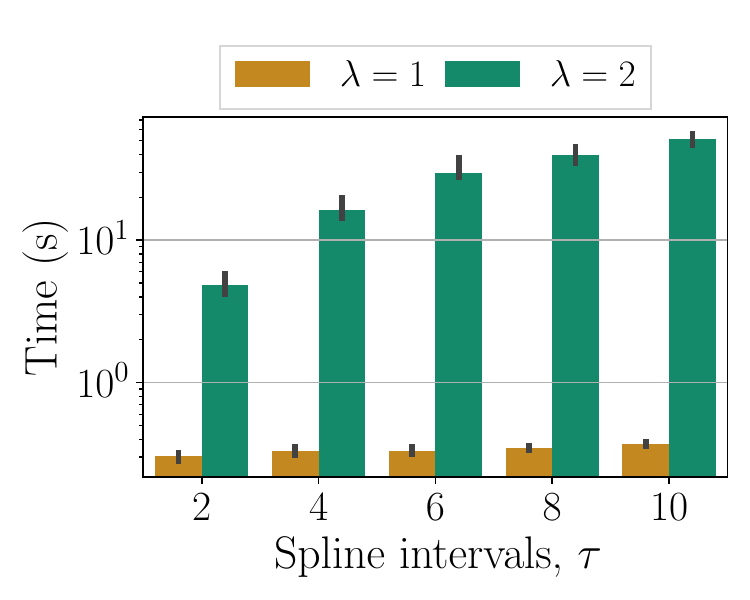}}
	
	\caption{Effect of spline intervals on the approximation error of statistical parity, root-mean square error (RMSE), and  execution time of {\framework}.}
	\label{fig:effect_spline_intervals_appendix}
\end{figure}

\subsection{Ablation Study: Effect of Spline Intervals} 
To understand the impact of spline intervals $\tau$ on {\framework}, we conduct an experiment. $\tau$ determines the number of local points to include in the cubic-spline based smoothing, with higher values providing better approximation of the component functions in the set-additive decomposition of the classifier (ref.\ Eq.~\eqref{eq:set_additive_classifier}). As shown in Figure~\ref{fig:effect_spline_intervals_appendix}, as $\tau$ increases, the approximation error of statistical parity based on FIFs decreases as well as the root mean square error of the set-additive approximation of the classifier. On the other hand, with higher $ \tau $, the execution time of {\framework} increases. Therefore, \emph{$ \tau $ exhibits a trade-off between the estimation accuracy and execution time of {\framework}.}

\begin{figure}
	\subfloat{\includegraphics[scale=0.4]{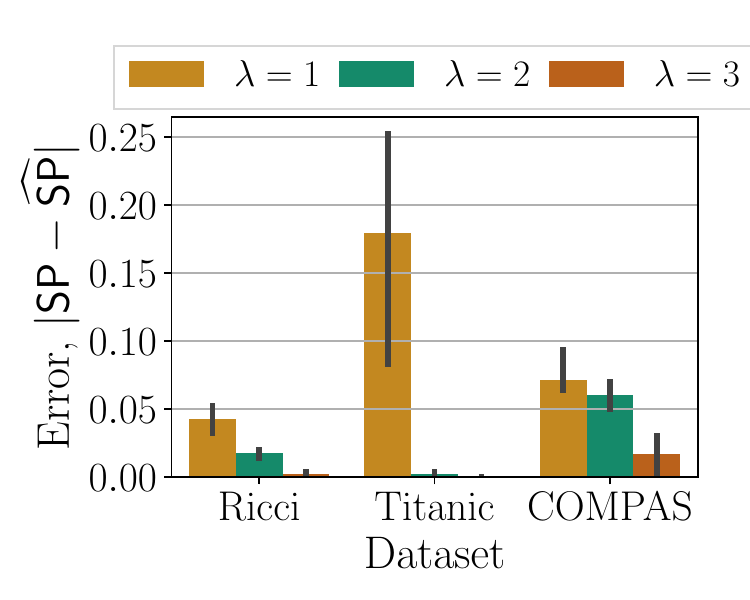}}
	\subfloat{\includegraphics[scale=0.4]{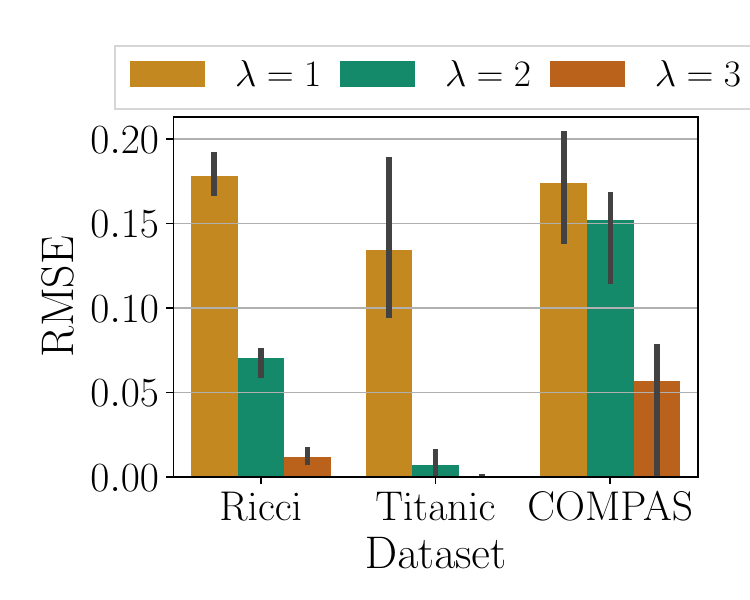}}
	\subfloat{\includegraphics[scale=0.4]{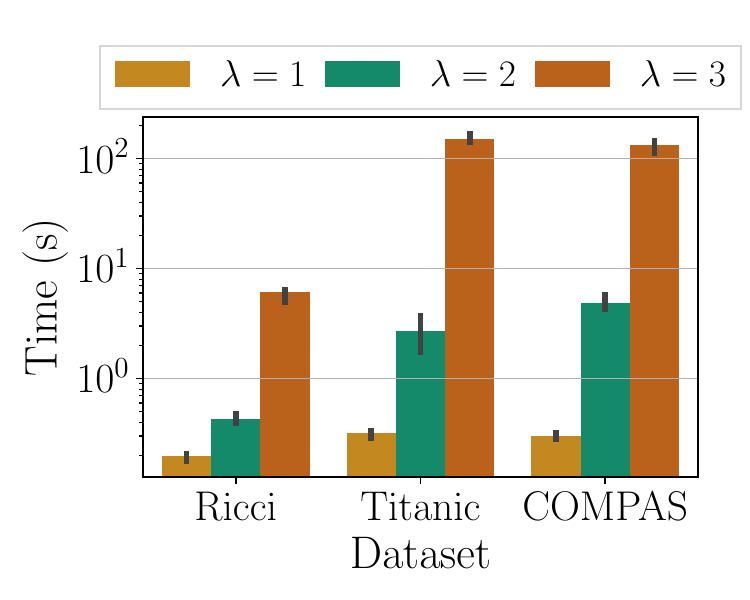}}

	\caption{Effect of maximum order $ \lambda $ on the approximation error of statistical parity, root-mean square error (RMSE) and  execution time of {\framework}.}
	\label{fig:effect_maxorder_appendix}
\end{figure}

\subsection{Ablation Study: Effect of Maximum Order of Intersectionality}\label{sec:ablation} 
Figure~\ref{fig:effect_maxorder_appendix} examines the impact of the maximum order of intersectionality ($\lambda$) on {\framework} in terms of accuracy and execution time on Ricci~\cite{mcginley2010ricci}, Titanic, and COMPAS datasets. As $\lambda$ increases, we see a decrease in approximation error for statistical parity based on FIFs, a decrease in the root mean squared error of the classifier's set additive decomposition, and an increase in execution time across different datasets. This means $\lambda$ provides a trade-off between accuracy and efficiency in {\framework}.

\subsection{FIF of Different Datasets}
\begin{figure}[!t]
	\begin{minipage}{0.45\textwidth}
		\centering
		\subfloat[Individual FIFs]{\includegraphics[scale=0.4]{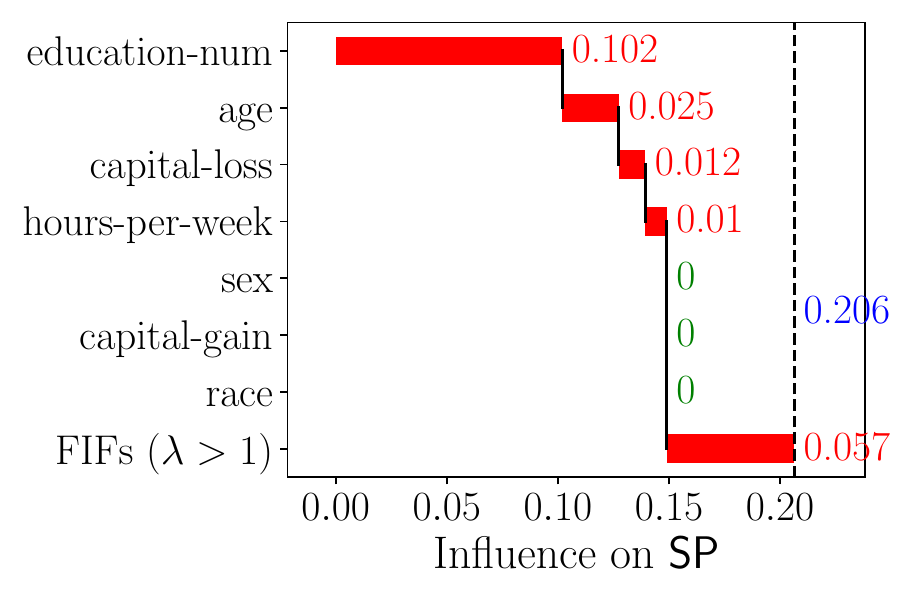}\label{fig:individual_fifs_adult}}
	\end{minipage}
	\begin{minipage}{0.5\textwidth}
		\centering
		\subfloat[Individual and intersectional FIFs]{\includegraphics[scale=0.4]{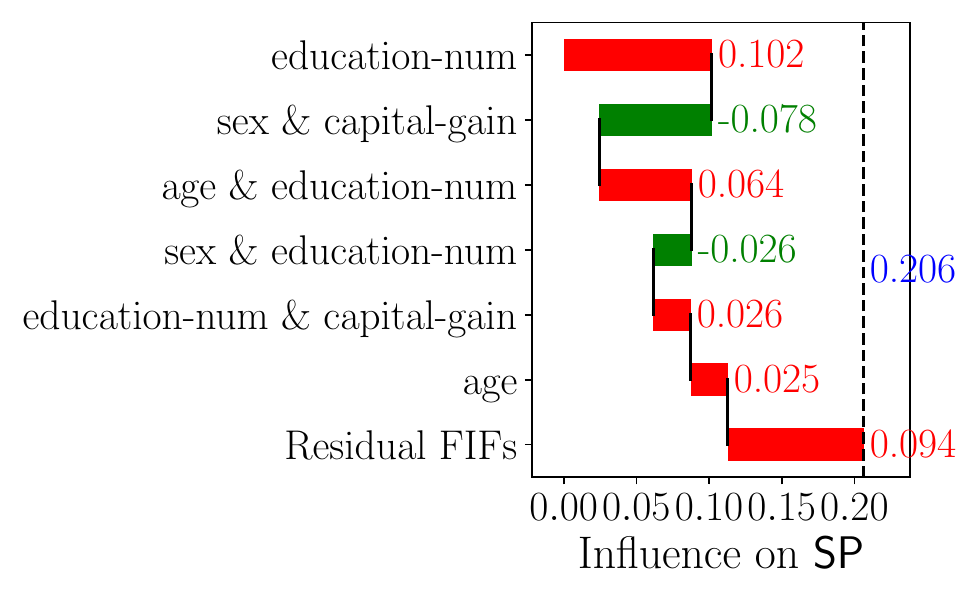}\label{fig:individual_and_intersectional_fifs_adult}}
	\end{minipage}
	\vspace{-0.2em}
	\caption{FIFs for Adult dataset on explaining statistical parity.}\label{fig:individual_vs_intersectional_influence_adult}
\end{figure}

\begin{figure}
	\centering
	\subfloat[Individual FIFs]{\includegraphics[scale=0.4]{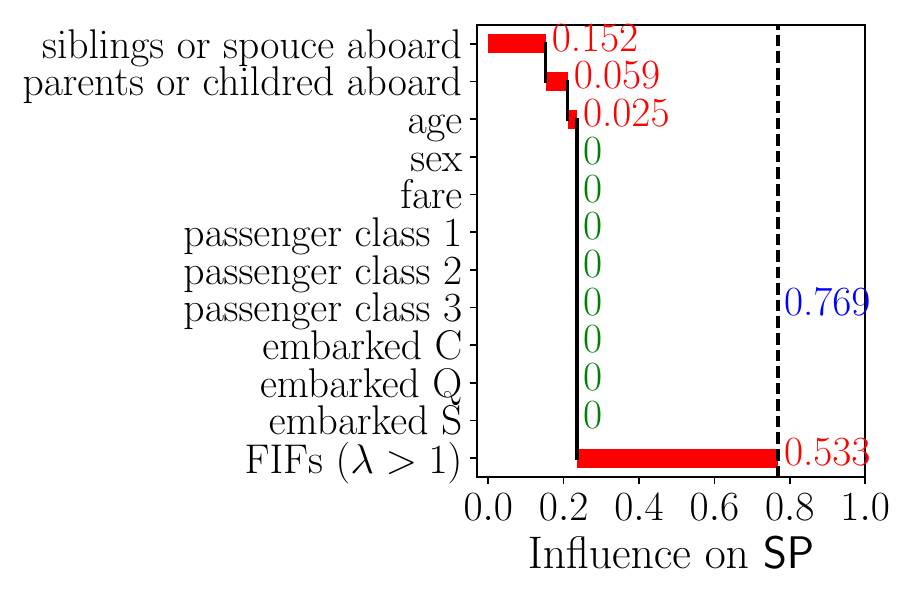}\label{fig:individual_fifs_titanic}}\hfil
	\subfloat[Individual and intersectional FIFs]{\includegraphics[scale=0.4]{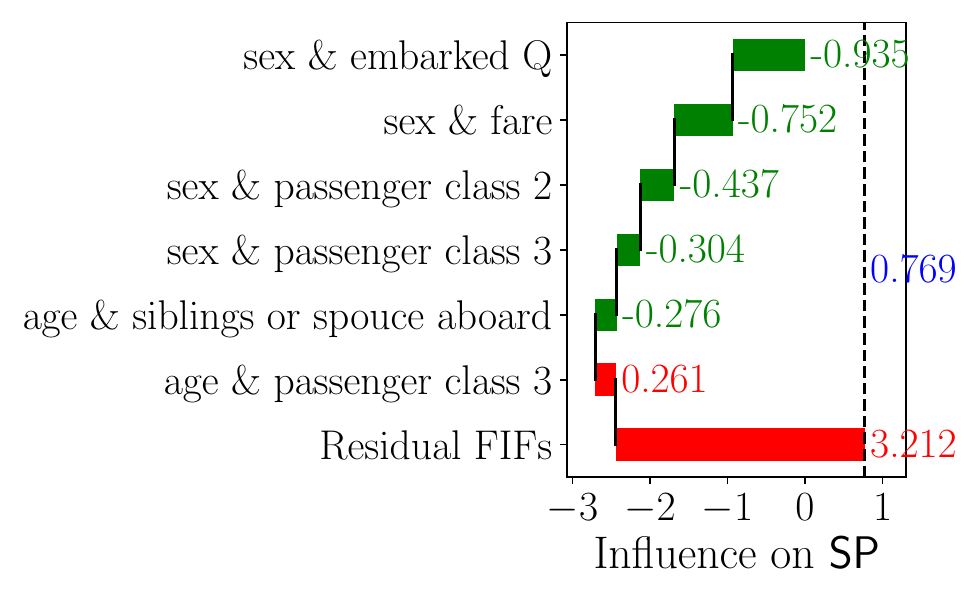}\label{fig:individual_and_intersectional_fifs_titanic}}
	\caption{FIFs for Titanic dataset on explaining statistical parity.}\label{fig:individual_vs_intersectional_influence_titanic}
\end{figure}

We deploy a neural network ($ 3 $ hidden layers, each with $ 2 $ neurons, L$ 2 $ penalty regularization term as $ 10^{-5} $, a constant learning rate as $ 0.001 $) on different datasets, namely Adult and Titanic, and demonstrate the corresponding FIFs in Figures~\ref{fig:individual_vs_intersectional_influence_adult} and \ref{fig:individual_vs_intersectional_influence_titanic}, respectively. In all the figures, both individual and intersectional FIFs depict the sources of bias more clearly than individual FIFs alone, as argued in Section~\ref{sec:experiments}.

\textbf{In Adult dataset,} the classifier predicts whether an individual earns more than $ \$50 $k per year or not, where race and sex are sensitive features. We observe that the trained network is unfair and it demonstrates statistical parity as $ 0.23 $. As we analyze FIFs,  education number, age, and capital gain/loss are key features responsible for the bias. 

\textbf{In Titanic dataset,} the neural network predicts whether a person survives the Titanic shipwreck or not. In this experiment, we consider the sex of a person as a sensitive feature and observe that the classifier is highly unfair achieving statistical parity as $ 0.83 $. Our FIF analysis reveals high correlation in Titanic, where individual FIFs are mostly zero while intersectional FIFs achieve high absolute values.

\end{document}